\documentclass{article}

\usepackage{enumitem}
\usepackage{forest}
\usepackage{listings}
\usepackage{subcaption}
\usepackage{soul}
\usepackage{tikz}
\usetikzlibrary{automata,positioning,arrows}
\usepackage{pgf-umlsd}

\begin{document}

\title{Parallel and in-process compilation of individuals for genetic programming on GPU}
\author{Hakan Ayral \\ hayral@gmail.com  \and Song\"{u}l Albayrak \\ songul@ce.yildiz.edu.tr}

\date{April 2017}
\maketitle

\begin{abstract}

Three approaches to implement genetic programming on GPU hardware are compilation, interpretation and direct generation of machine code. The compiled approach is known to have a prohibitive overhead compared to other two.

This paper investigates methods to accelerate compilation of individuals for genetic programming on GPU hardware. We apply in-process compilation to minimize the compilation overhead at each generation; and we investigate ways to parallelize in-process compilation. In-process compilation doesn't lend itself to trivial parallelization with threads; we propose a multiprocess parallelization using memory sharing and operating systems interprocess communication primitives. With parallelized compilation we achieve further reductions on compilation overhead. Another contribution of this work is the code framework we built in C\# for the experiments. The framework makes it possible to build arbitrary grammatical genetic programming experiments that run on GPU with minimal extra coding effort, and is available as open source.
\end{abstract}

\section*{Introduction}

Genetic programming is an evolutionary computation technique, where the objective is to find a program (i.e. a simple expression, a sequence of statements, or a full-scale function) that satisfy a behavioral specification expressed as test cases along with expected results. Grammatical genetic programming is a subfield of genetic programming, where the search space is restricted to a language defined as a BNF grammar, thus ensuring all individuals to be syntactically valid.

Processing power provided by graphic processing units (GPUs) make them an attractive platform for evolutionary computation due to the inherently parallelizable nature of the latter. First genetic programming implementations shown to run on GPUs were \cite{adataparallelapproach} and \cite{fastgeneticprogrammingongpus}.

Just like in the CPU case, genetic programming on GPU requires the code represented by individuals to be rendered to an executable form; this can be achieved by compilation to an executable binary object, by conversion to an intermediate representation of a custom interpreter developed to run on GPU, or by directly generating machine-code for the GPU architecture. Compilation of individuals' codes for GPU is known to have a prohibitive overhead that is hard to offset with the gains from the GPU acceleration.

Compiled approach for genetic programming on GPU is especially important for grammatical genetic programming; the representation of individuals for linear and cartesian genetic programming are inherently suitable for simple interpreters and circuit simulators implementable on a GPU. On the other hand grammatical genetic programming aims to make higher level constructs and structures representable, using individuals that represent strings of tokens belonging to a language defined by a grammar; unfortunately executing such a representation sooner or later requires some form of compilation or complex interpretation.

In this paper we first present three benchmark problems we implemented to measure compilation times with. We use grammatical genetic programming for the experiments, therefore we define the benchmark problems with their grammars, test cases and fitness functions.

Then we set a baseline by measuring the compilation time of individuals for those three problems, using the conventional CUDA compiler Nvcc. Afterwards we measure the speedup obtained by the in-process compilation using the same benchmark problem setups. We proceed by presenting the obstacles encountered on parallelization of in-process compilation. Finally we propose a parallelization scheme for in-process compilation, and measure the extra speedup achieved.

\section*{Prior Work}  %  -------------------------------------------------

\cite{distributedgeneticprogramming} deals with the compilation overhead of individuals for genetic programming on GPU using CUDA. Article proposes a distributed compilation scheme where a cluster of around 16 computers compile different individuals in parallel; and states the need for large number of fitness cases to offset the compilation overhead. It correctly predicts that this mismatch will get worse with increasing number of cores on GPUs, but also states that "a large number of classic benchmark GP problems fit into this category". Based on figure 5 of the article it can be computed that for a population size of 256, authors required \textit{25 ms/individual} in total\footnote{This number includes network traffic, XO, mutation and processing time on GPU, in addition to compilation times. In our case the difference between compilation time and total time has constantly been at sub-millisecond level per population on all problems; thus for comparison purposes compile times we present can also be taken as total time with an error margin of $^{1ms}/_{pop. size}$}.

\cite{evolvingacudakernel} presents first use of grammatical genetic programming on the GPU, applied to a string matching problem to improve gzip compression; with a grammar constructed from fragments of an existing string matching CUDA code. Based on figure 11 of the accompanying technical report\cite{evolvingacudakernel-techreport} a population of 1000 individuals (10 kernels of 100 individuals each) takes around 50 seconds to compile using nvcc from CUDA v2.3 SDK, which puts the average compilation time to approximately \textit{50 ms/individual}.

In \cite{graphicsprocessingunitsandgeneticprogramming} an overview of genetic programming on GPU hardware is provided, along with a brief presentation and comparison of compiled and interpreted approaches. As part of the comparison it underlines the trade-off between the speed of compiled code versus the overhead of compilation, and states that the command line CUDA compiler was especially slow, hence why interpreted approach is usually preferred.

\cite{accelerationofgrammatical} investigate the acceleration of grammatical evolution by use of GPUs, by considering performance impact of different design decisions like thread/block granularity, different types of memory on GPU, host-device memory transactions. As part of the article compilation to PTX form and loading to GPU with JIT compilation on driver level, is compared with directly compiling to cubin object and loading to GPU without further JIT compilation. For a kernel containing 90 individuals takes 540ms to compile to CUBIN with sub-millisecond upload time to GPU, vs 450ms for compilation to PTX and 80ms for JIT compilation and upload to GPU using nvcc compiler from CUDA v3.2 SDK. Thus PTX+JIT case which is the faster of the two achieves average compilation time of \textit{5.88 ms/individual}.

\cite{identifyingsimilaritiesintmbl} proposes an approach for improving compilation times of individuals for genetic programming on GPU, where common statements on similar locations  are aligned as much as possible across individuals. After alignment individuals with overlaps are merged to common kernels such that aligned statements become a single statement, and diverging statements are enclosed with conditionals to make them part of the code path only if the value of individual\_ID parameter matches an individual having that divergent statements. Authors state that in exchange for faster compilation times, they get slightly slower GPU runtime with  merged kernels as all individuals need to evaluate every condition at the entry of each divergent code block coming from different individuals. In results it is stated that for individuals with 300 instructions, compile time is 347 ms/individual if it's unaligned, and \textit{72 ms/individual} if it's aligned (time for alignment itself not included) with nvcc compiler from CUDA v3.2 SDK.

\cite{evolvinggpumachinecode} provides a comparison of compilation, interpretation and direct generation of machine code methods for genetic programming on GPUs. Five benchmark problems consisting of Mexican Hat and Salutowicz regressions, Mackey-Glass time series forecast, Sobel Filter and 20-bit Multiplexer are used to measure the comparative speed of the three mentioned methods. It is stated that compilation method uses nvcc compiler from CUDA V5.5 SDK. Compilation time breakdown is only provided for Mexican Hat regression benchmark on Table 6, where it is stated that total nvcc compilation time took 135,027 seconds and total JIT compilation took 106,458 seconds. Table 5 states that Mexican Hat problem uses 400K generations and a population size of 36. Therefore we can say that an average compilation time of $^{(135,027 + 106,458)}/_{36\times 400,000} =$ \textit{16.76 ms/individual} is achieved.

%\section*{Methods}  %  ----------------------------------------------------

\section*{Implemented Problems for Measurement}  %  ----------------------------------------------------

We implemented three problems as benchmark to compare compilation speed. They consist of a general program synthesis problem, Keijzer-6 as a regression problem \cite{keijzer}, and 5-bit Multiplier as a multi output boolean problem . The latter two are included in the  "Alternatives to blacklisted problems" table on \cite{bettergpbenchmarks}.

We use grammatical genetic programming as our representation and phenotype production method; therefore all problems are defined with a BNF grammar that defines a search space of syntactically valid programs, along with some test cases and a fitness function specific to the problem. For all three problems, a genotype which is a list of (initially random) integers derives to a phenotype which is a valid CUDA C expression, or code block in form of a list of statements. All individuals are appended and prepended with some initialization and finalization code which serves to setup up the input state and write the output to GPU memory afterwards. See Appendix for BNF Grammars and codes used to surround the individuals.

\subsection*{Search Problem}

Search Problem is designed to evolve a function which can identify whether a search value is present in an integer list, and return its position if present or return -1 otherwise.

We first proposed this problem as a general program synthesis benchmark in \cite{effectsofpopulation}. The grammar for the problem is inspired by \cite{experimentsinprogramsynthesis}; we designed it to be a subproblem of the more general \textit{integer sort problem} case along with some others. It also bears some similarity to problems presented in \cite{generalprogramsynthesis} based on, the generality of its usecase, combined with simplicity of its implementation.

Test cases consist of unordered lists of random integers in the range $[0,50]$, and list lengths vary between 3 and 20. Test cases are randomly generated but half of them are ensured to contain the value searched, and others ensured not to contain. We employed a binary fitness function, which returns 1 if the returned result is correct (position of searched value or -1 if it's not present on list) or 0 if it's not correct; hence the fitness of an individual is the sum of its fitnesses over all test cases, which evolutionary engine tries to maximize.

\subsection*{Keijzer-6}

Keijzer-6 function, introduced in \cite{keijzer}, is the function $K_6(x) = \sum_{n=1}^x \frac{1}{n}$ which maps a single integer parameter to the partial sum of harmonic series with number of terms indicated by its parameter. Regression of Keijzer-6 function is one of the recommended alternatives to replace simpler symbolic regression problems like quartic polynomial \cite{bettergpbenchmarks}.

For this problem we used a modified version of the grammar given in \cite{managingrepetition}, and \cite{exploringpositionindependent}, with the only modification of increasing constant and variable token ratio as the expression nesting gets deeper. We used the root mean squared error as fitness function which is the accepted practice for this problem.

\subsection*{5-bit multiplier}
5-bit multiplier problem consists of finding a boolean relation that takes 10 binary inputs to 10 binary outputs, where two groups of 5 inputs each represent an integer up to $2^5-1$ in binary, and the output represents a single integer up to $2^{10}-1$, such that the output is the multiplication of the two input numbers. This problem is generally attacked as 10 independent binary regression problems, with each bit of the output is separately evolved as a circuit or boolean function.

It's easy to show that the number of $n$-bit input $m$-bit output binary relations are ${2^{m(2^n)}}$, which grows super-exponentially. Multiple output multiplier is the recommended alternative to Multiplexer and Parity problems in \cite{bettergpbenchmarks}

We transfer input to and output from GPU with bits packed as a single 32bit integer; hence there is a code preamble before first individual to unpack the input bits, and a post-amble after each individual to pack the 10 bits computed by evolved expressions as an integer.

The fitness function for 5-bit multiplier computes the number of bits different between the individual's response and correct answer, by computing the pop count of these two XORed.

\section*{Development and Experiment Setup}  %  ----------------------------------------------------

\subsection*{Hardware Platform}

All experiments have been conducted on a dual Xeon E5-2670 (8 physical 16 logical cores per CPU, 32 cores in total) platform running at 2.6Ghz equipped with 60GB RAM, along with dual SSD storage and four NVidia GRID K520 GPUs. Each GPU itself consists of 1536 cores spread through 8 multiprocessors running at 800Mhz, along with 4GB GDDR5 RAM \footnote{see validation of hardware used at experiment: http://www.techpowerup.com/gpuz/details/7u5xd/} and is able to sustain 2 teraflops of single precision operations (\textit{in total 6144 cores and 16GB GDDR5 VRAM which can theoretically sustain 8 teraflops single precision computation assuming no other bottlenecks}). GPUs are accessed for computation through NVidia CUDA v8 API and libraries, running on top of Windows Server 2012 R2 operating system.

\subsection*{Development Environment}

Codes related to grammar generation, parsing, derivation, genetic programming, evolution, fitness computation and GPU access has been implemented in C\#, using \textit{managedCuda} \footnote{{https://kunzmi.github.io/managedCuda/}} for CUDA API bindings and NVRTC interface, along with \textit{CUDAfy.NET} \footnote{{https://cudafy.codeplex.com/}} for interfacing to NVCC command line compiler. The grammars for the problems has been prepared such that the languages defined are valid subsets of CUDA C language specialized towards the respective problems.

\subsection*{Experiment Parameters}

We ran each experiment with population sizes starting from 20 individual per population, going up to 300 with increments of 20. As the subject of interest is compilation times and not fitness, we measured the following three parameters to evaluate compilation speed:
\begin{enumerate}[label=(\roman*)]
\item \textit{ptx} : Cuda source code to Ptx compilation time per individual
\item \textit{jit} : Ptx to Cubin object compilation time per individual
\item \textit{other} : All remaining operations a GP cycle requires (i.e compiled individuals running on GPU, downloading produced results, computing fitness values, evolutionary selection, cross over, mutation, etc.)
\end{enumerate}

The value of \textit{other} is measured to be always at sub-millisecond level, in all experiments, all problems and for all population sizes. Therefore it does not appear on plots. For all practical purposes $ptx+jit$ can be considered as the total time cost of a complete cycle for a generation, with an error margin of $\frac{1 ms}{pop. size}$.

Each data point on plots corresponds to the average of one of those measurements for the corresponding $(population size, measurement type, experiment)$ triple. Each average is computed over the measurement values obtained for the first 10 generations of 15 different populations for given size (thus effectively the compile times of 150 generations averaged). The reason for not using 150 generations of a single population directly is that a population gains bias towards to a certain type of individuals after certain number of generations, and stops representing the inherent unbiased distribution of grammar.

The number of test cases used is dependent to the nature of problem; on the other hand as each test case is run as a GPU thread, it is desirable that the number of test cases are a multiple of 32 on any problem, as finest granularity for task scheduling on modern GPUs is a group of 32 threads which is called a \textit{Warp}. For non multiple of 32 test cases, GPU transparently rounds up the number to nearest multiple of 32 and allocate cores accordingly, with some threads from the last warp work on cores with output disabled. The number of test cases we used  during experiments were 32 for \textit{Search Problem}, 64 for regression of Keijzer-6 function and 1024 ($=2^{(5+5)}$) for 5-bit Binary Multiplier Problem. For all experiments both mutation and crossover rate was set to 0.7; these rates do not affect the compilation times.

\section*{Experiment Results}  %  -----------------------------------------------

\subsection*{Conventional Compilation as Baseline}

\begin{figure}[!htb]

    \begin{subfigure}[b]{0.5\textwidth}
            \centering
            \includegraphics[width=\textwidth]{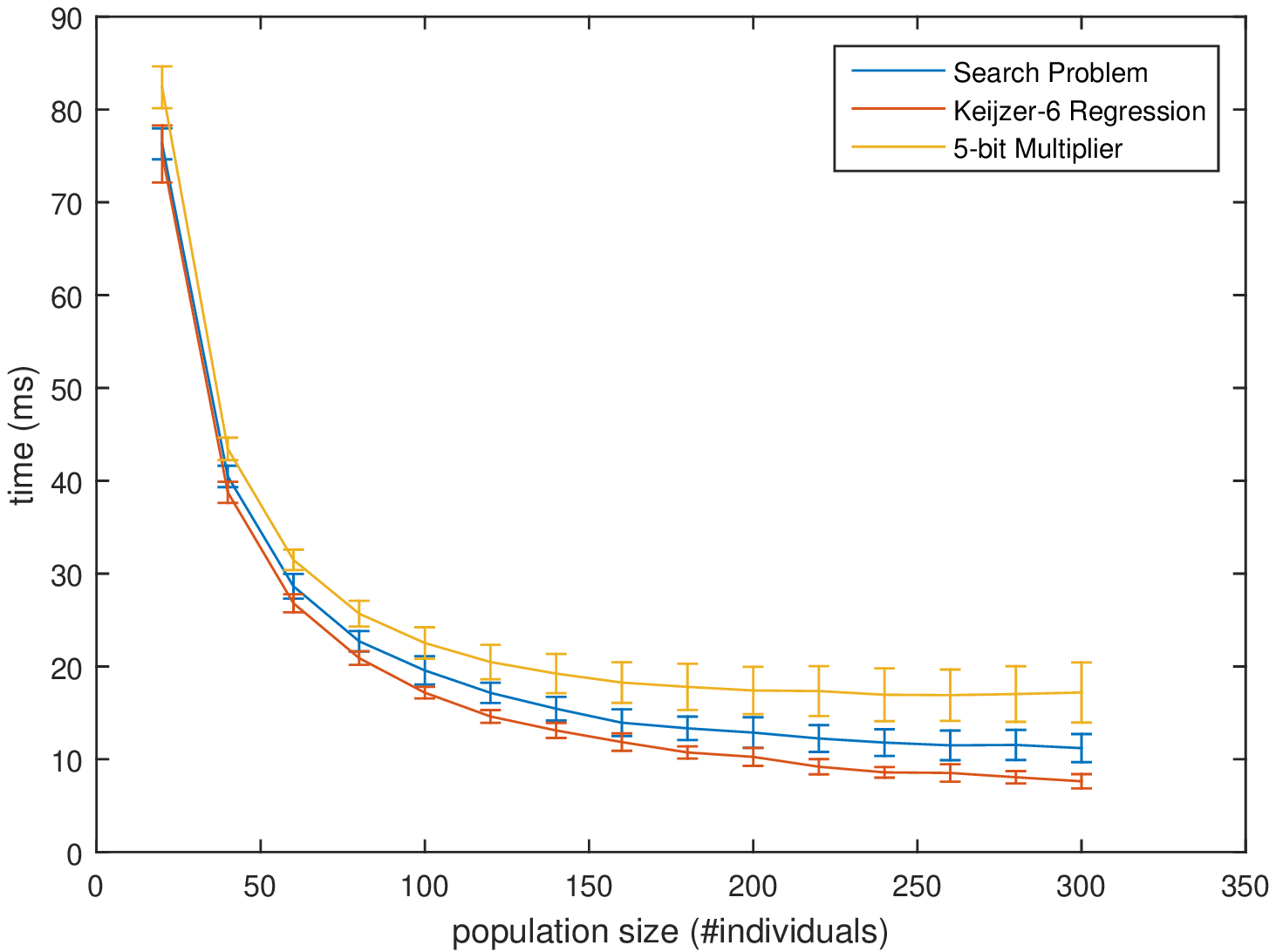}
    \caption{Per individual compile time}
    \end{subfigure}
\begin{subfigure}[b]{0.5\textwidth}
            \centering
            \includegraphics[width=\textwidth]{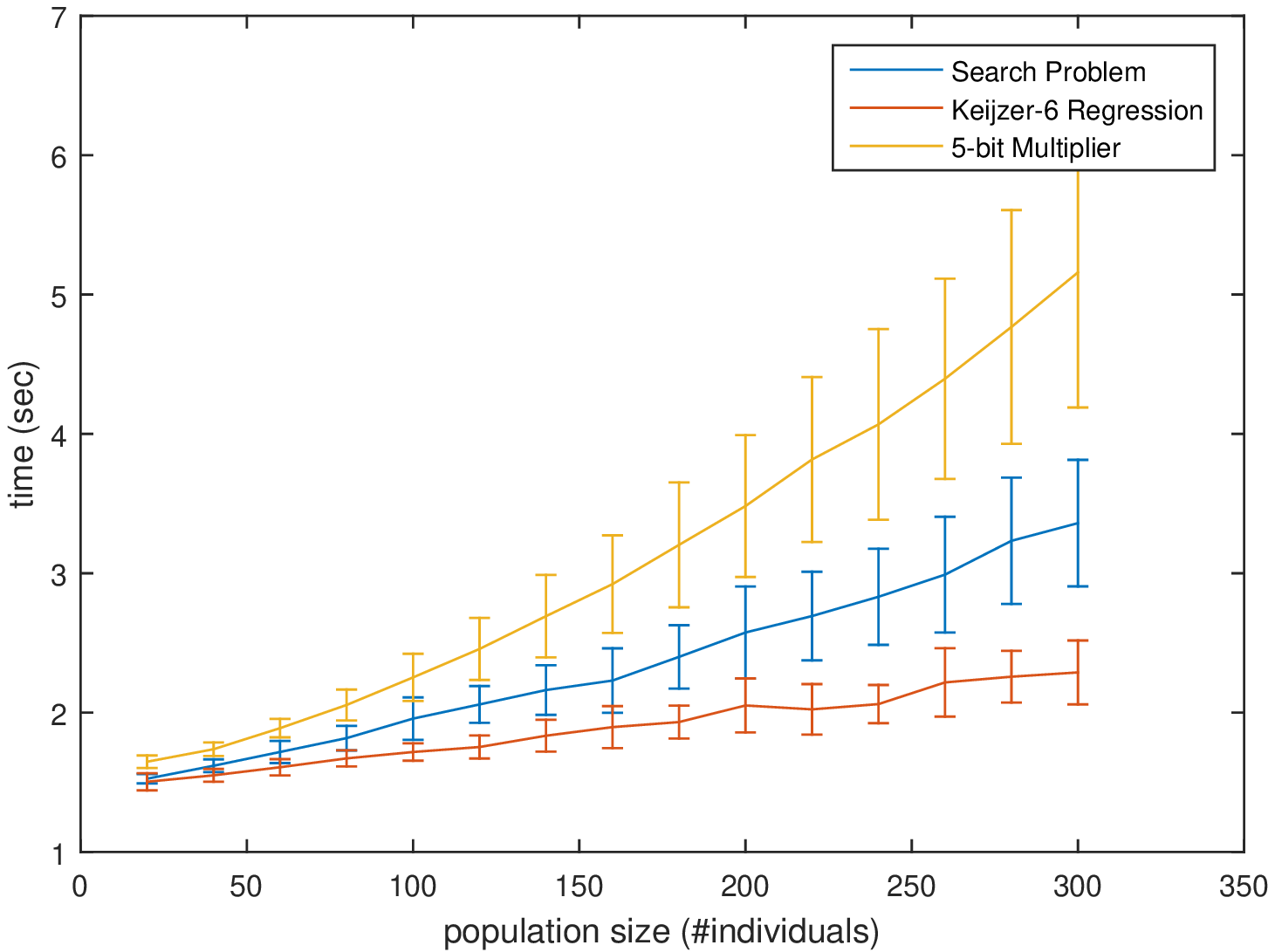}
    \caption{Total compile time}
    \end{subfigure}

\caption{Nvcc compilation times by population size.}% (left) Per individual (right) Total}
\label{fig:all-three-nvcc}
\end{figure}

NVCC is the default compiler of CUDA platform, it is distributed as a command line application. In addition to compilation of cuda C source codes, it performs tasks such as the separation of source code as host code and device code, calling the underlying host compiler (GCC or Visual C compiler) for host part of source code, and linking compiled host and device object files.

Fib\ref{fig:all-three-nvcc}(a) shows that compilation times level out at 11.2 ms/individual for Search Problem, at 7.62 ms/individual for Keijzer-6 regression, and at 17.2 ms/individual for 5-bit multiplier problem. It can be seen on Fig.\ref{fig:all-three-nvcc}(b) that, even though not obvious, the total compilation time does not increase linearly, which is most observable on trace of 5-bit multiplier problem. As Nvcc is a separate process, it isn't possible to measure the distribution of compilation time between source to ptx, ptx to cubin, and all other setup work (i.e. process launch overhead, disk I/O); therefore it is not possible to pinpoint the source of nonlinearity on total compilation time.

The need for successive invocations of Nvcc application, and all data transfers being handled over disk files are the main drawbacks of Nvcc use in a real time\footnote{not as in hard real time, but as prolonged, successive and throughput sensitive use} context, which is the case in genetic programming. Eventhough the repeated creation and teardown of NVCC process most probably guarantees that the application stays on disk cache, this still prevents it to stay cached on processor L1/L2 caches.

\subsection*{In-process Compilation}
NVRTC is a runtime compilation library for CUDA C, it was first released as part of v7 of CUDA platform in 2015. NVRTC accepts CUDA source code and compiles it to PTX in-memory. The PTX string generated by NVRTC can be further compiled to device dependent CUBIN object file and loaded with CUDA Driver API still without persisting it to a disk file. This provides optimizations and performance not possible in off-line static compilation.

Without NVRTC, for each compilation a separate process needs to be spawned to execute nvcc at runtime. This has significant overhead drawback, NVRTC addresses these issues by providing a library interface that eliminates overhead of spawning separate processes, and extra disk I/O.

\begin{figure}[!htb]

    \begin{subfigure}[b]{0.5\textwidth}
            
            \includegraphics[width=\textwidth]{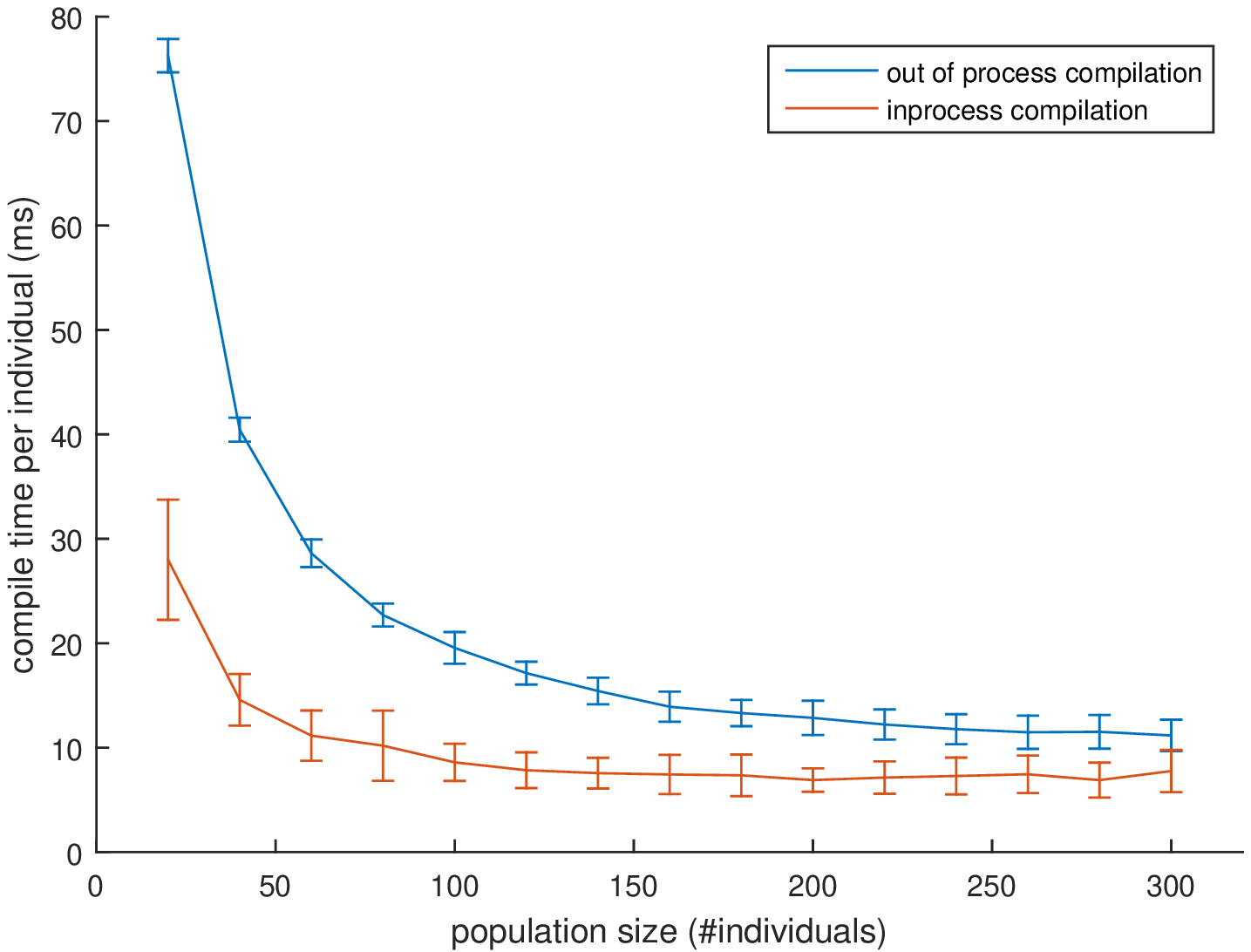}
    \caption{ Per individual}
    \end{subfigure}
	\begin{subfigure}[b]{0.5\textwidth}
            
            \includegraphics[width=\textwidth]{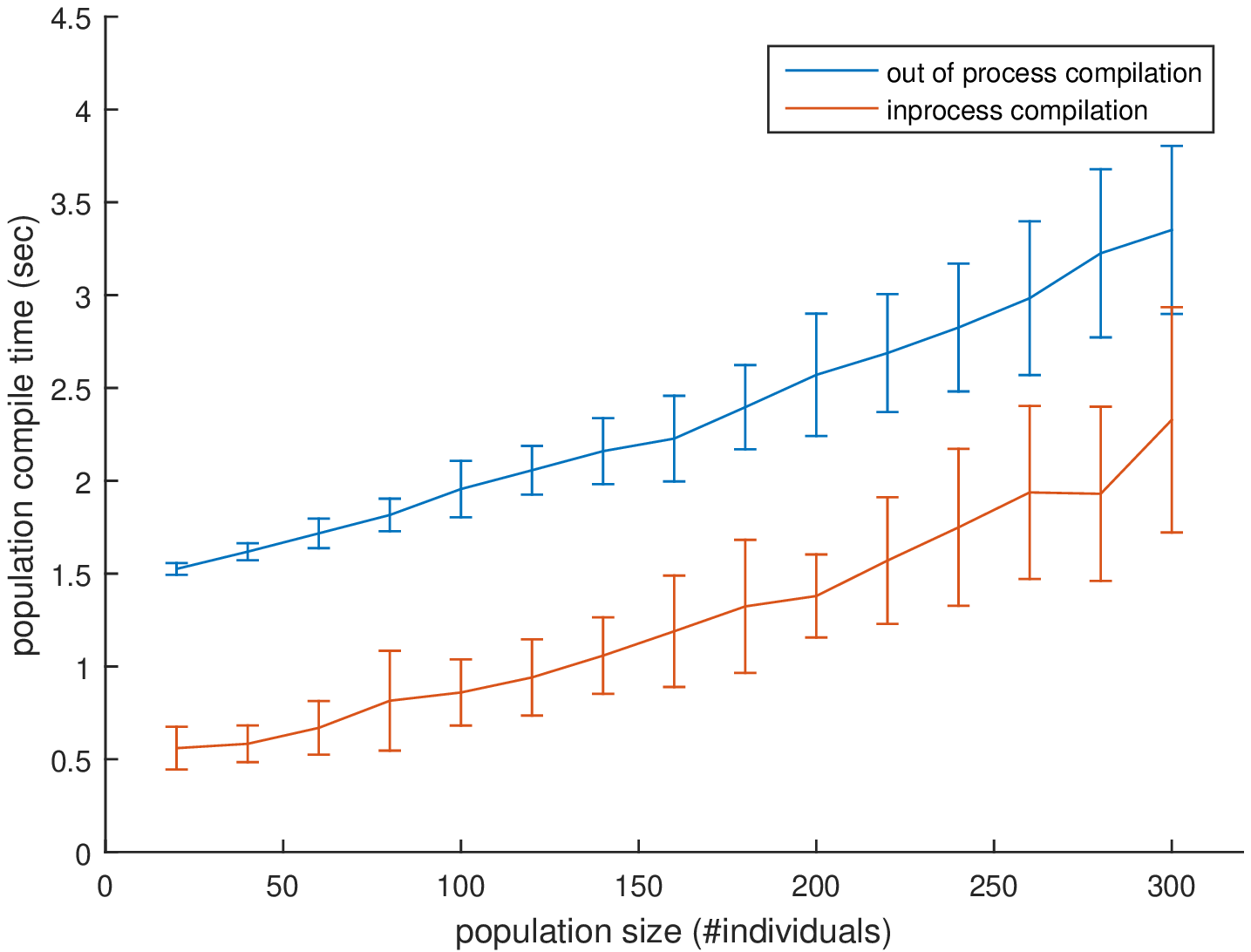}
    \caption{ Total}
    \end{subfigure}
\caption{In-process and out of process compilation times by population size, for Search Problem} % (left) Per individual (right) Total}
\label{fig:search-nvcc-vs-nvrtc}
\end{figure}
\begin{figure}[!htb]
\begin{subfigure}[b]{0.5\textwidth}
            
            \includegraphics[width=\textwidth]{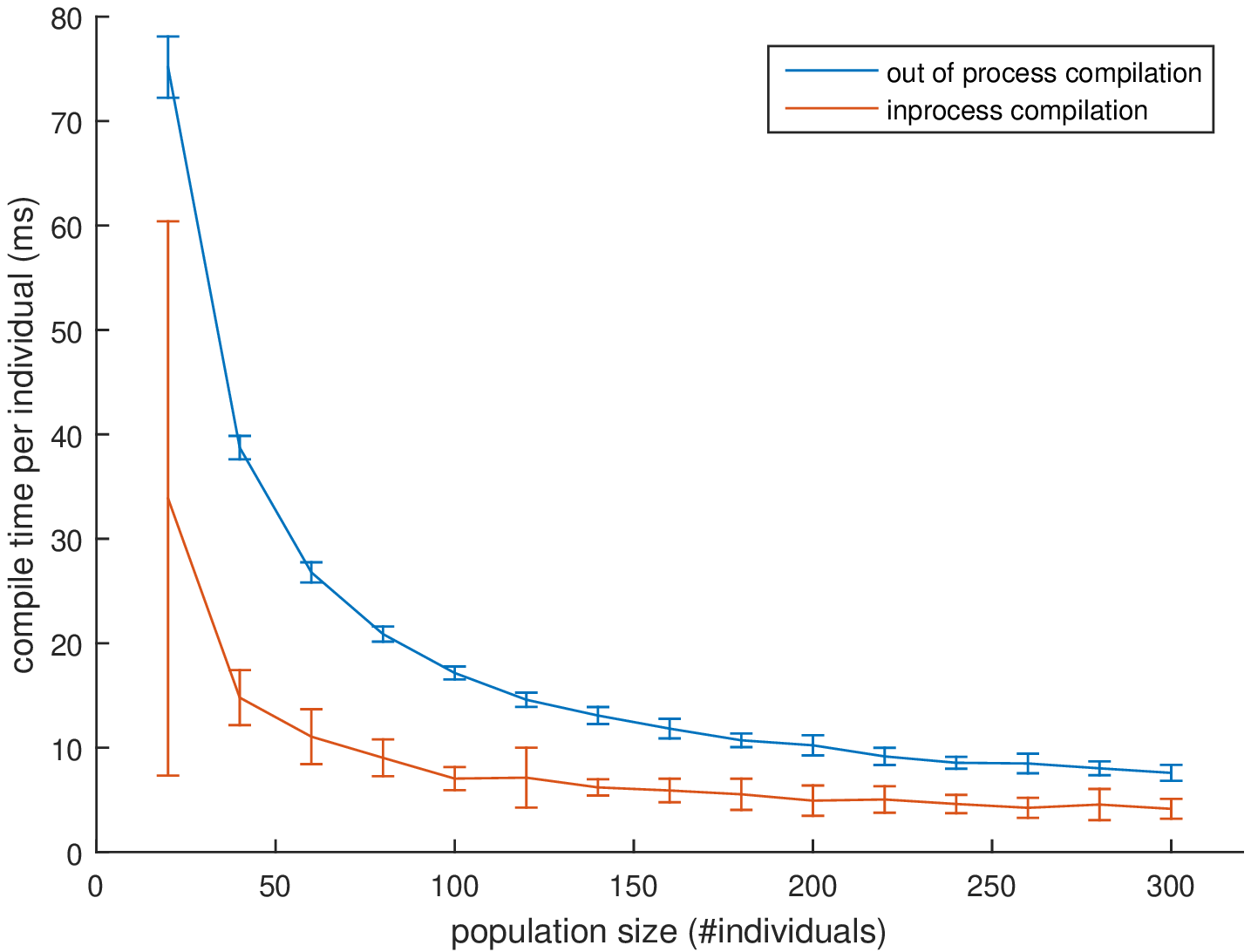}
    \caption{ Per individual}
    \end{subfigure}
	\begin{subfigure}[b]{0.5\textwidth}
            
            \includegraphics[width=\textwidth]{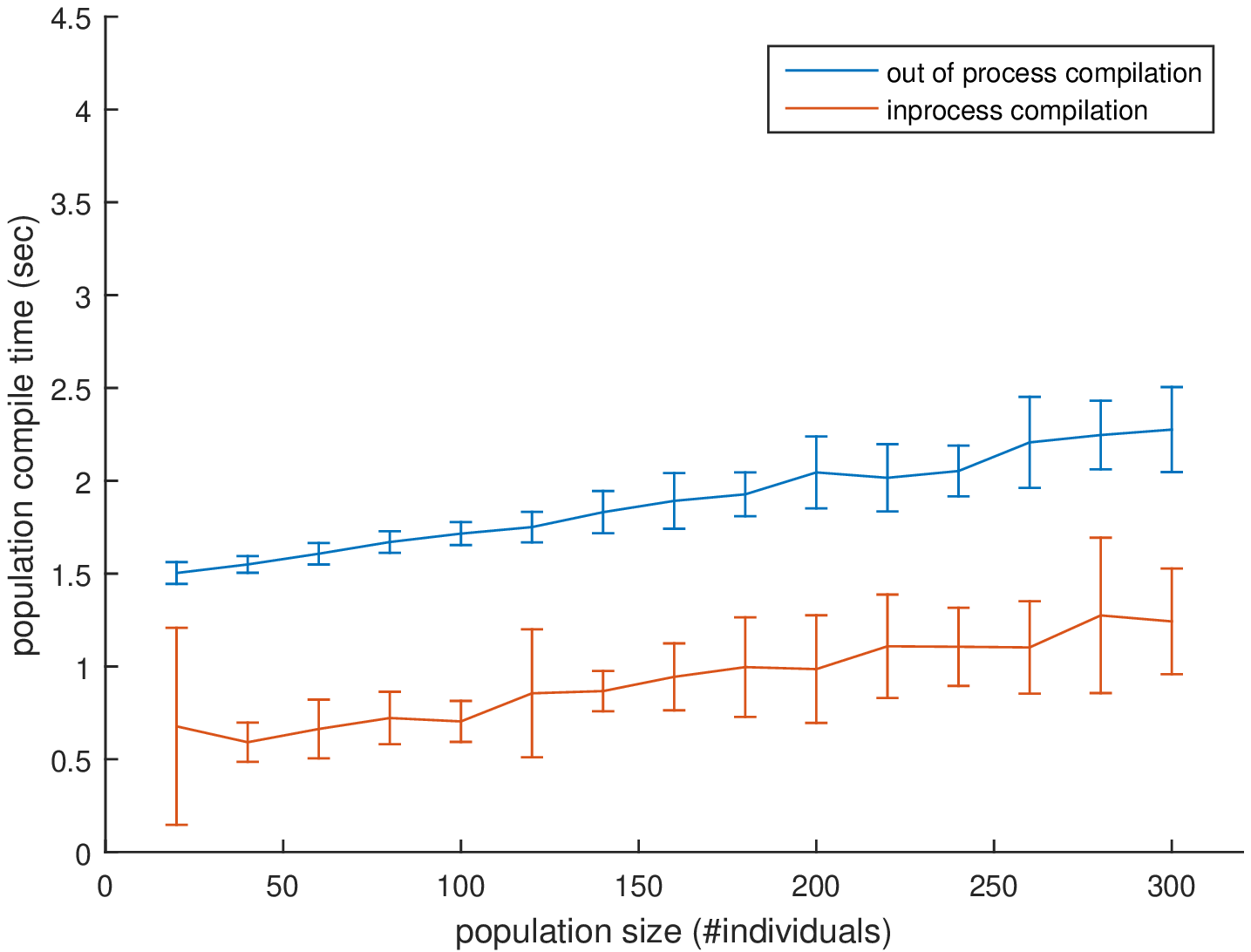}
    \caption{ Total}
    \end{subfigure}
\caption{In-process and out of process compilation times by population size, for Keijzer-6 Regression} % (left) Per individual (right) Total}
\label{fig:k6-nvcc-vs-nvrtc}
\end{figure}

\begin{figure}[!htb]
    \begin{subfigure}[b]{0.5\textwidth}
            
            \includegraphics[width=\textwidth]{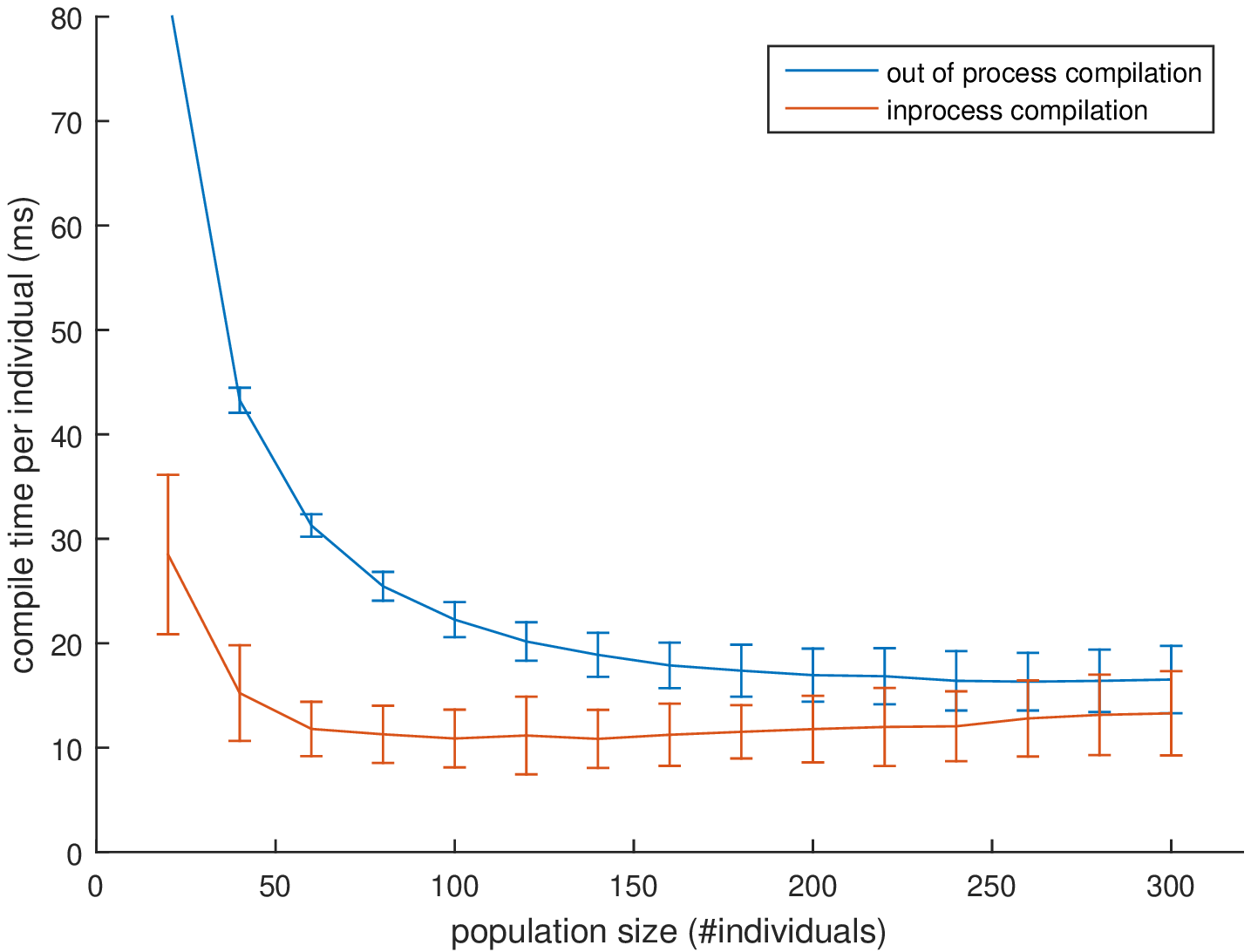}
    \caption{Per individual}
    \end{subfigure}
	\begin{subfigure}[b]{0.5\textwidth}
            
            \includegraphics[width=\textwidth]{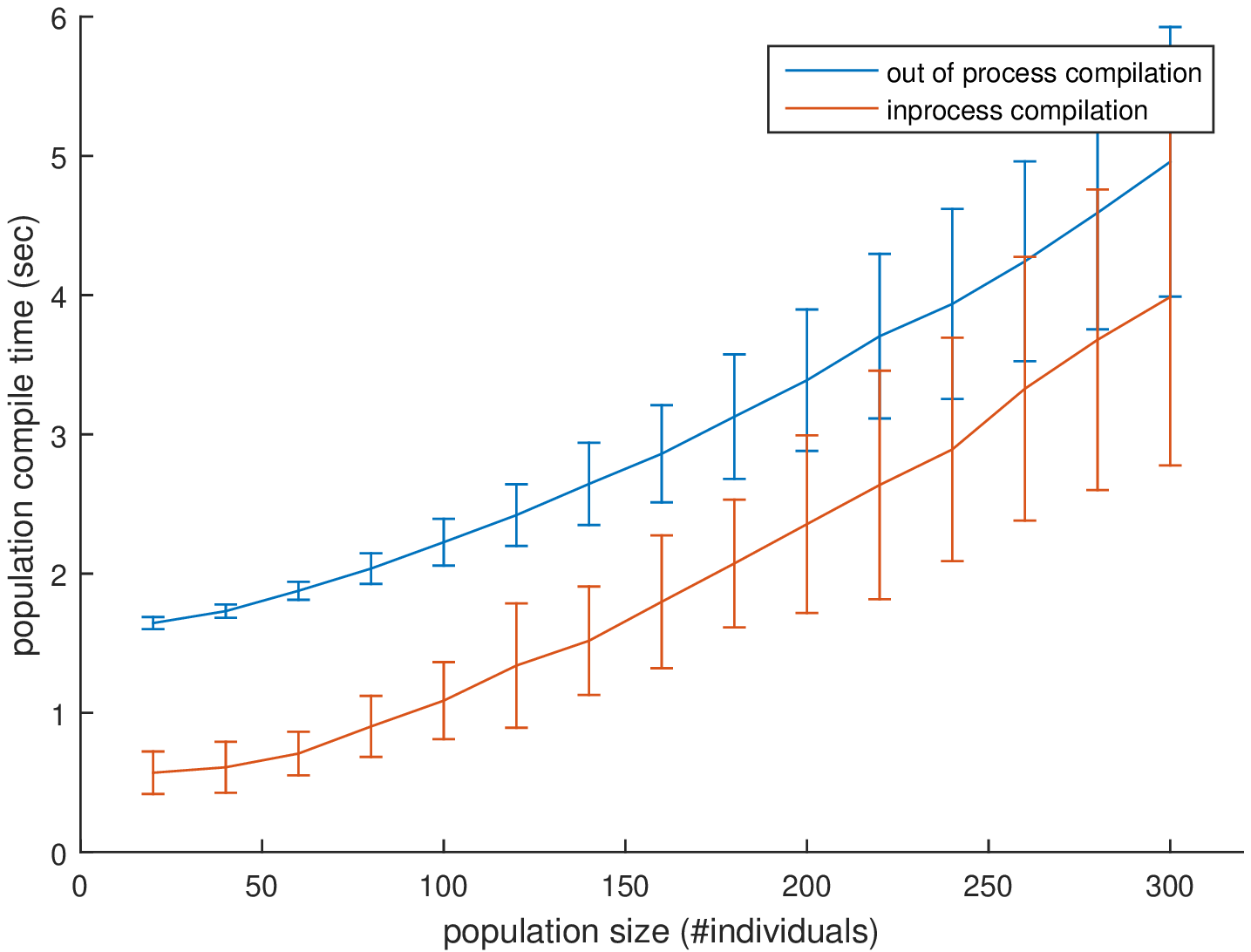}
    \caption{ Total}
    \end{subfigure}

\caption{In-process and out of process compilation times by population size, for 5-bit Multiplier} % (left) Per individual (right) Total}
\label{fig:mul-nvcc-vs-nvrtc}
\end{figure}
\begin{figure}[!htb]
\centering
\includegraphics[width=0.6\textwidth]{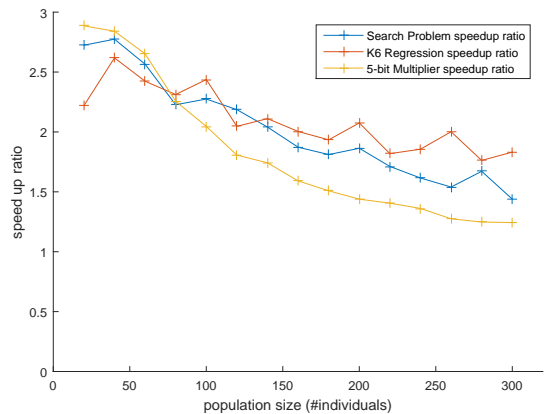}

\caption{Compile time speedup ratios between conventional and in-process compilation by problem}
\label{fig:all-speedups}
\end{figure}

On figures \ref{fig:search-nvcc-vs-nvrtc},\ref{fig:k6-nvcc-vs-nvrtc} and \ref{fig:mul-nvcc-vs-nvrtc} it can be seen that in-process compilation of individuals not only provides reduced compilation times for all problems on all population sizes, it also allows to reach asymptotically optimal per individual compilation time with much smaller populations.
The fastest compilation times achieved with in-process compilation is 4.14 ms/individual for Keijzer-6 regression (at 300 individuals per population), 10.88 ms/individual for 5-bit multiplier problem (at 100 individuals per population\footnote{compilation speed at 300 individuals per population is 13.29 ms/individual}), and 6.89 ms/individual for Search Problem (at 280 individuals per population\footnote{compilation speed at 300 individuals per population is 7.76 ms/individual}).
The total compilation time speed ups are measured to be in the order of 261\% to 176\% for the K6 regression problem, 288\% to 124\% for the 5-bit multiplier problem, and 272\% to 143\% for the Search Problem, depending on population size (see Fig.\ref{fig:all-speedups}).

\subsection*{Parallelizing In-process Compilation}

\subsubsection*{Infeasibility of parallelization with threads}

A first approach to parallelize in-process compilation, comes to mind as to partition the individuals and spawn multiple threads that will compile each partition in parallel through NVRTC library. Unfortunately it turns out that NVRTC library is not designed for multi-threaded use; we noticed that when multiple compilation calls are made from different threads at the same time, the execution is automatically serialized.

\begin{figure}[!htb]
\center
\includegraphics[width=\textwidth]{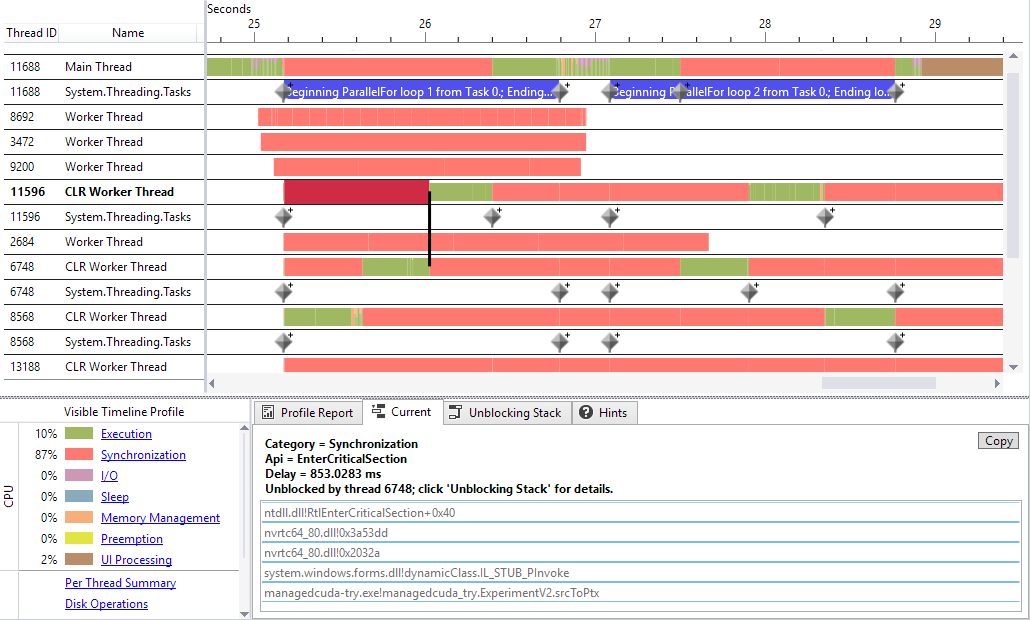}
\caption{NVRTC library serializes calls from multiple threads}
\label{fig:nvrtc-serialized}
\end{figure}

Stack trace in Fig.\ref{fig:nvrtc-serialized} shows \textit{nvrtc64\_80.dll} calling OS kernel's \textit{EnterCriticalSection} function to block for exclusive execution of a code block, and gets unblocked by another thread which also runs a block from same library, 853ms later via the release of the related lock. The pattern of green blocks on three threads in addition to main thread in Fig.\ref{fig:nvrtc-serialized} shows that calls are perfectly serialized one after another, despite being called at the same time which is hinted by the red synchronization blocks preceding them.

Although NVRTC compiles CUDA source to PTX with a single call, the presence of compiler options setup function which affects the following compilation call, and use of critical sections at function entries, show that apparently this is a stateful API. Furthermore, unlike CUDA APIs' design, mentioned state is most likely not stored in thread local storage (TLS), but stored on the private heap of the dynamic loading library, making it impossible for us to trivially parallelize this closed source library using threads, as moving the kept state to TLS requires source level modifications.

\subsubsection*{Parallelization with daemon processes}

Therefore as a second approach we implemented a daemon process which stays resident. It is launched from command line with a unique ID as command line parameter to allow multiple instances. Instances of daemon is launched as many times as the wanted level of parallelism, and each instance identifies itself with the ID received as parameter. Each launched process register two named synchronization events with the operating system, for signaling the state transitions of a simple state machine consisting of $\{starting,available,processing  \}$ states which represent the state of that instance. Main process also has copies of same state machines for each instance to track the states of daemons. Thus both processes (main and daemon) keep a consistent view of the mirrored state machine by monitoring the named events which allows state transitions to be performed in lock step. State transition can be initiated by both processes, specifically $(starting \to available)$ and $(processing \to available)$ is triggered by the daemon, and $(available \to processing)$ is triggered by the main process.

\begin{figure}[!htb]
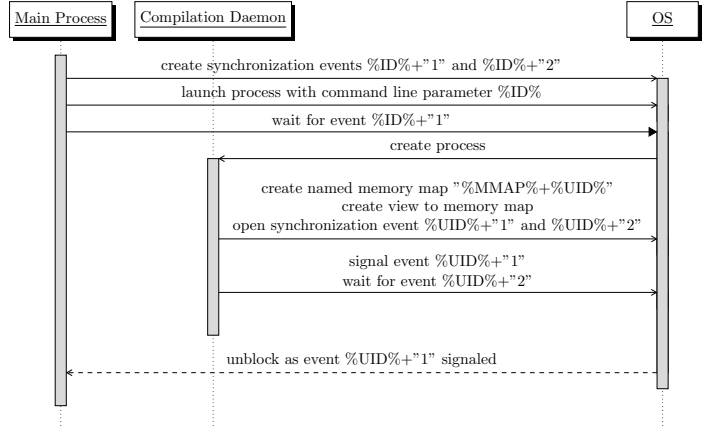

\centering
\resizebox{0.8\textwidth}{!}{
  \begin{sequencediagram}
    \newthread{A}{Main Process}{}
    \newinst[0.5]{B}{Compilation Daemon}{}
    \newinst[7.5]{OS}{OS}{}
    \begin{messcall}{A}{create synchronization events \%ID\%+"1" and \%ID\%+"2"}{OS} 
    \begin{messcall}{A}{launch process with command line parameter \%ID\%}{OS} \end{messcall}
    \prelevel
    \begin{call}{A}{wait for event \%ID\%+"1"}{OS}{unblock as event \%UID\%+"1" signaled} 
      \begin{messcall}{OS}{create process}{B} 
      		\postlevel \postlevel
          \begin{messcall}{B}{ \shortstack{  create named memory map "\%MMAP\%+\%UID\%" \\ create view to memory map \\ open synchronization event \%UID\%+"1" and \%UID\%+"2"}}{OS} \end{messcall}
          \begin{messcall}{B}{ \shortstack{ signal event \%UID\%+"1" \\ wait for event \%UID\%+"2" }}{OS} \end{messcall}
        \end{messcall}
     \end{call}
     
     \end{messcall}
  \end{sequencediagram}
}  
  
  \caption{Sequence Diagram for creation of a compilation daemon process and related interprocess communication primitives}
  \label{fig:sequence1}
\end{figure}

The communication between the main process and compilation daemons are handled via shared views to memory maps. Each daemon register a named memory map and create a memory view, onto which main process also creates a view to after the daemon signals state transition from $starting$ to $available$. (see Fig.\ref{fig:sequence1}) CUDA source is passed through this shared memory, and compiled device dependent CUBIN object file is also returned through the same. To signal the state transition $(starting \to available)$ daemon process signals the first event and starts waiting for the second event at the same time. Once a daemon leaves the $starting$ state, it never returns back to it.

\begin{figure}[!htb]
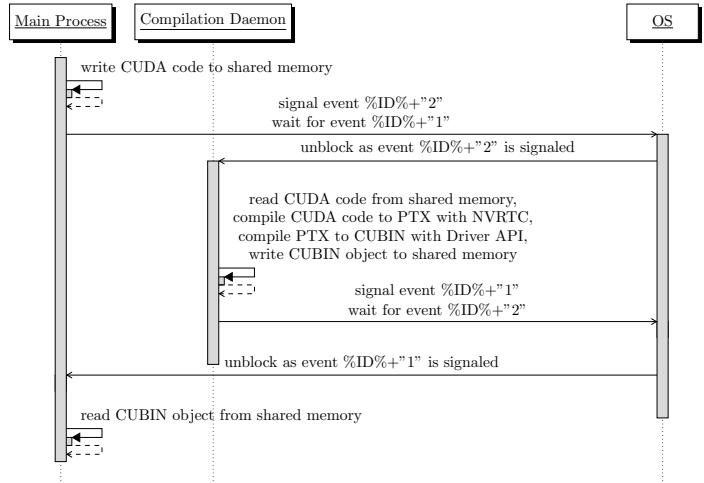

\centering
\resizebox{0.8\textwidth}{!}{
  \begin{sequencediagram}
    \newthread{A}{Main Process}{}
    \newinst[0.5]{B}{Compilation Daemon}{}
    \newinst[7.5]{OS}{OS}{}
    
    \begin{call}{A}{write CUDA code to shared memory}{A}{} \end{call}
    
    \begin{messcall}{A}{\shortstack{signal event \%ID\%+"2" \\ wait for event \%ID\%+"1"}}{OS} 
    
    	\begin{messcall}{OS}{unblock as event \%ID\%+"2" is signaled}{B}
        \postlevel \postlevel \postlevel
          \begin{call}{B}{\shortstack{read CUDA code from shared memory,\\compile CUDA code to PTX with NVRTC, \\ compile PTX to CUBIN with Driver API,\\write CUBIN object to shared memory}}{B}{}
          
          \end{call}

          \begin{messcall}{B}{\shortstack{signal event \%ID\%+"1" \\ wait for event \%ID\%+"2" } }{OS} \end{messcall}                
          
        \end{messcall}
        \prelevel
        \begin{messcall}{OS}{unblock as event \%ID\%+"1" is signaled}{A} \end{messcall}   
    \end{messcall}
    \prelevel
    \begin{call}{A}{read CUBIN object from shared memory}{A}{} \end{call}
  \end{sequencediagram}
}
  \caption{Sequence Diagram for compilation on daemon process and related interprocess communication}
  \label{fig:sequence2}
\end{figure}

When the main process generate a new population to be compiled it partitions the individuals in a balanced way, such that the difference of number of individuals between any pair of partitions is never more than one. Once the individuals are partitioned, the generated CUDA codes for each partition are passed to the daemon processes. Each daemon waits in the blocked state till main process wakes that specific daemon for a new batch of source to compile by signaling the second named event of that process (see Fig.\ref{fig:sequence2}). Main process signals all daemons asynchronously to start compiling; then starts waiting for the completion of daemon processes' work. To prevent the UI thread of main process getting blocked too, main process maintains a separate thread for each daemon process it communicates with, therefore while waiting for daemon processes to finish their jobs only those threads of main process are blocked. Main process signaling the second event and daemon process unblocking as a result, corresponds to the state transition $(available \to processing)$.

When a daemon process arrives to $processing$ state, it reads the CUDA source code from the shared view of the memory map related to its ID, and compiles the code using NVRTC library. 

Once a daemon finishes compiling and writes the Cubin object to shared memory, it signals the first event to unblock the related thread in main process and starts to wait for the second event once again. This signaling, blocking pair corresponds to the state transition $(processing \to available)$.

\subsubsection*{Cost of Parallelization}

The parallelization approach we propose is virtually overhead free when compared to a hypothetical parallelization scenario using threads. As the daemon processes are already resident and waiting in the memory along with the loaded NVRTC library, the overhead of both parallelization approaches is limited to the time cost of memory moves from/to shared memory and synchronization by named events\footnote{on Windows operating system named events is the fastest IPC primitive, upon which all others (i.e. mutex, semaphore) are implemented}. The only difference between the two is, in a context switch between threads of same process, processor keeps the Translation Look Aside Buffer (TLB), but in case of a context switch to another process TLB is flushed as processor transitions to a new virtual address space; we conjecture that the impact would be negligible.

About the memory cost, all modern operating systems recognize when an executable binary or shared library gets loaded multiple times; OS keeps a single copy of the related memory pages on physical memory, and separately maps those to virtual address spaces of each process using those. This not only saves physical RAM, but also allows better space locality for L2/L3 processor caches. Hence memory consumption by multiple instances of our daemon processes each loading NVRTC library (\textit{nvrtc64\_80.dll} is almost 15MB) to their own address space, is almost the same as the consumption of a single instance.

\subsubsection*{Speedup Achieved with Parallel Compilation}

 \begin{table}[]
\centering
\caption{Compilation Times by Compilation Methods for Search Problem with 300 individuals}
\label{search-table}
\begin{tabular}{|c|r|r|r|r|} \hline
           & \multicolumn{2}{c|}{Compilation Time}       & \multicolumn{2}{c|}{Speedup ratio}   \\
     Compilation          &     &   & In-process   & Nvcc   \\ 
 Method & Per individual & Total &   compilation &   compilation \\ \hline
Nvcc        &  11.20 ms              & 3.36 sec       & -                         & 1.00                             \\
In-process  &  7.76 ms              & 2.33 sec      & 1.00                         & 1.44                              \\
2 daemons  &  3.81 ms              & 1.14 sec      & 2.04                          &  2.93                             \\
4 daemons  &  2.53 ms              & 0.76 sec      & 3.07                          &  4.41                             \\
6 daemons  &  2.23 ms              & 0.67 sec      & 3.48                          &  5.01                             \\
8 daemons  &  2.13 ms              & 0.64 sec      & 3.65                          &  5.26                             \\   \hline
\end{tabular}
\end{table}

\begin{table}[]  %  -----------------------------------------------------------
\centering
\caption{Compilation Times by Compilation Methods for Keijzer-6 Regression with 300 individuals}
\label{K6-table}
\begin{tabular}{|c|r|r|r|r|} \hline
           & \multicolumn{2}{c|}{Compilation Time}       & \multicolumn{2}{c|}{Speedup ratio}   \\
     Compilation          &     &   & In-process   & Nvcc   \\ 
 Method & Per individual & Total &   compilation &   compilation \\ \hline
Nvcc        & 7.63 ms               & 2.29 sec      & -                         & 1.00                             \\
In-process  & 4.14 ms               & 1.24 sec       & 1.00                         & 1.83                                  \\
2 daemons  &  2.92 ms              & 0.88 sec      & 1.42                              & 2.60                                   \\
4 daemons  &  2.45 ms              & 0.73 sec      & 1.69                              & 3.10                                  \\
6 daemons  &  2.20 ms              & 0.66 sec      & 1.88                              & 3.45                                  \\
8 daemons  &  2.25 ms              & 0.67 sec      & 1.84                              & 3.37                                  \\   \hline
\end{tabular}
\end{table}

\begin{table}[]  %  -----------------------------------------------------------
\centering
\caption{Compilation Times by Compilation Methods for 5-bit Multiplier Problem with 300 individuals}
\label{MUL-table}
\begin{tabular}{|c|r|r|r|r|} \hline
           & \multicolumn{2}{c|}{Compilation Time}       & \multicolumn{2}{c|}{Speedup ratio}   \\
     Compilation          &     &   & In-process   & Nvcc   \\ 
 Method & Per individual & Total &   compilation &   compilation \\ \hline
Nvcc        & 17.20 ms               & 5.16 sec       & -                         & 1.00                             \\
In-process  & 13.29 ms               & 3.99 sec       & 1.00                         & 1.24                                  \\
2 daemons  &   6.15 ms             &  1.85 sec     & 2.16                              & 2.69                                  \\
4 daemons  &   3.23 ms             &  0.97 sec     & 4.12                              & 5.12                                  \\
6 daemons  &   2.42 ms             &  0.73 sec     & 5.49                              & 6.82                                  \\
8 daemons  &   2.17 ms             &  0.65 sec     & 6.11                              & 7.60                                  \\   \hline
\end{tabular}
\end{table}

\begin{figure}[!ht]  %  ----------------

    \begin{subfigure}[b]{0.5\textwidth}
            \centering
            \includegraphics[width=\textwidth]{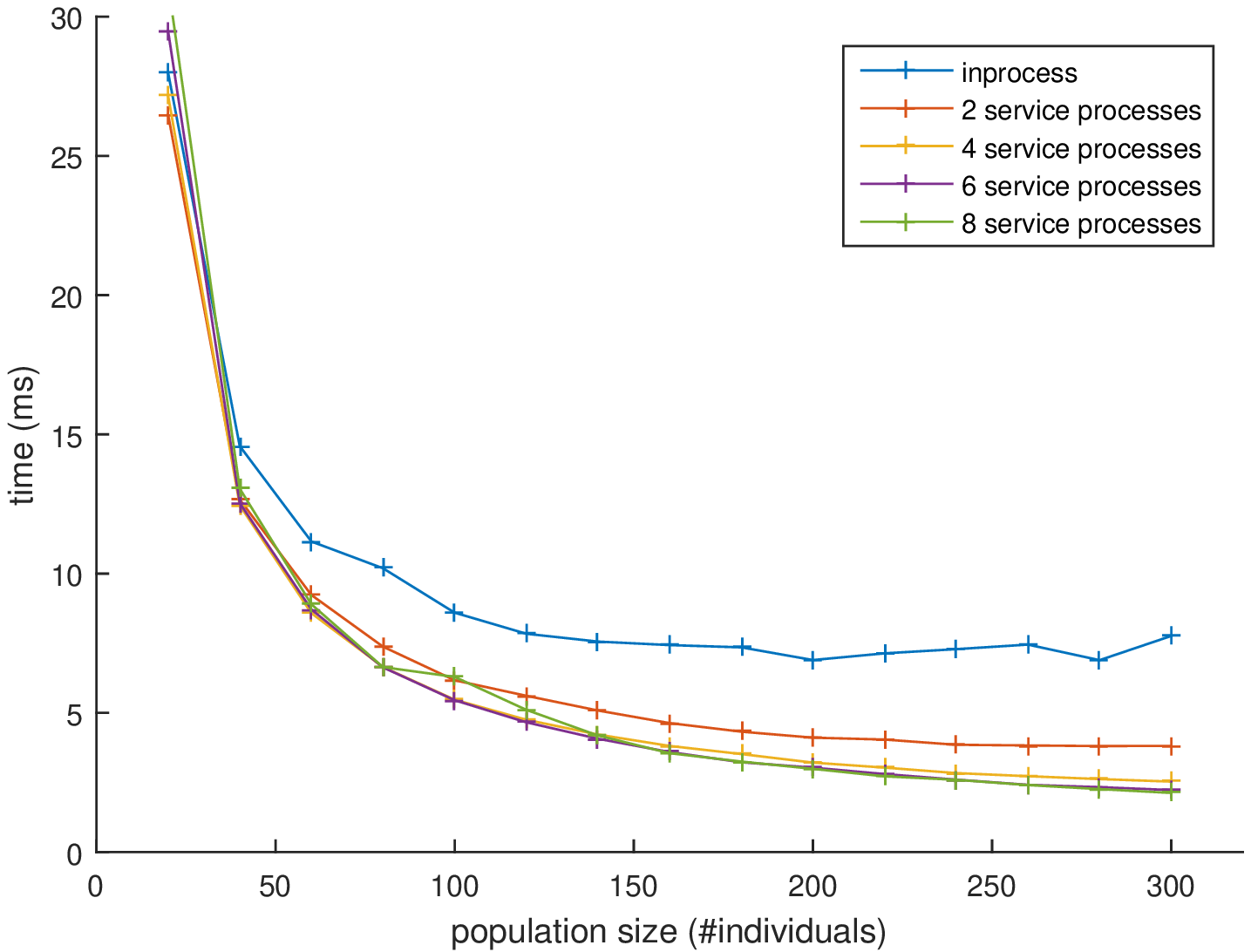}
    \caption{Per individual compile time}
    \end{subfigure}
\begin{subfigure}[b]{0.5\textwidth}
            \centering
            \includegraphics[width=\textwidth]{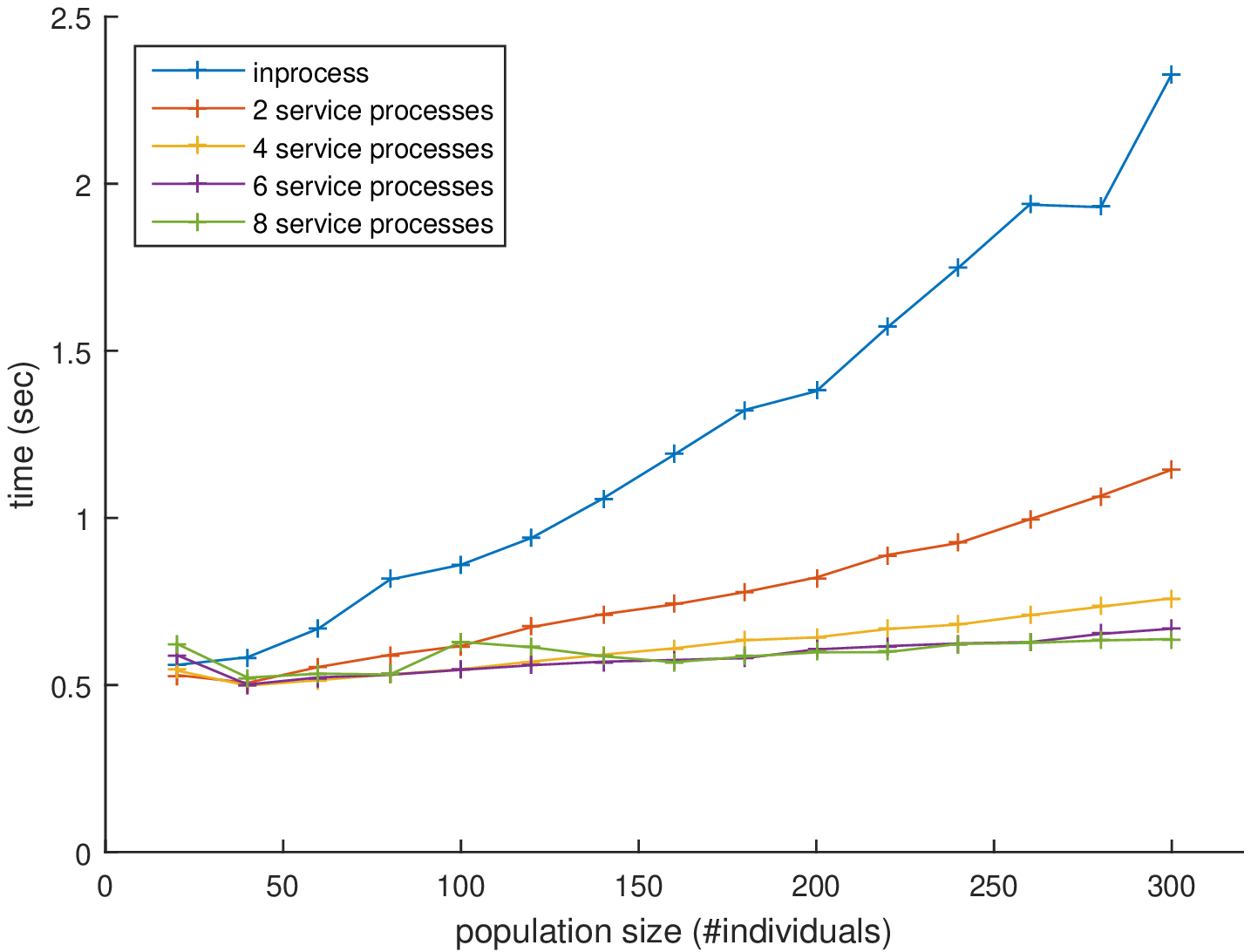}
    \caption{Total compile time}
    \end{subfigure}

\caption{Nvcc compilation times for Search Problem by number of servicing resident processes}
\label{fig:search-parallel}
\end{figure}

\begin{figure}[!ht]  %  ----------------

    \begin{subfigure}[b]{0.5\textwidth}
            \centering
            \includegraphics[width=\textwidth]{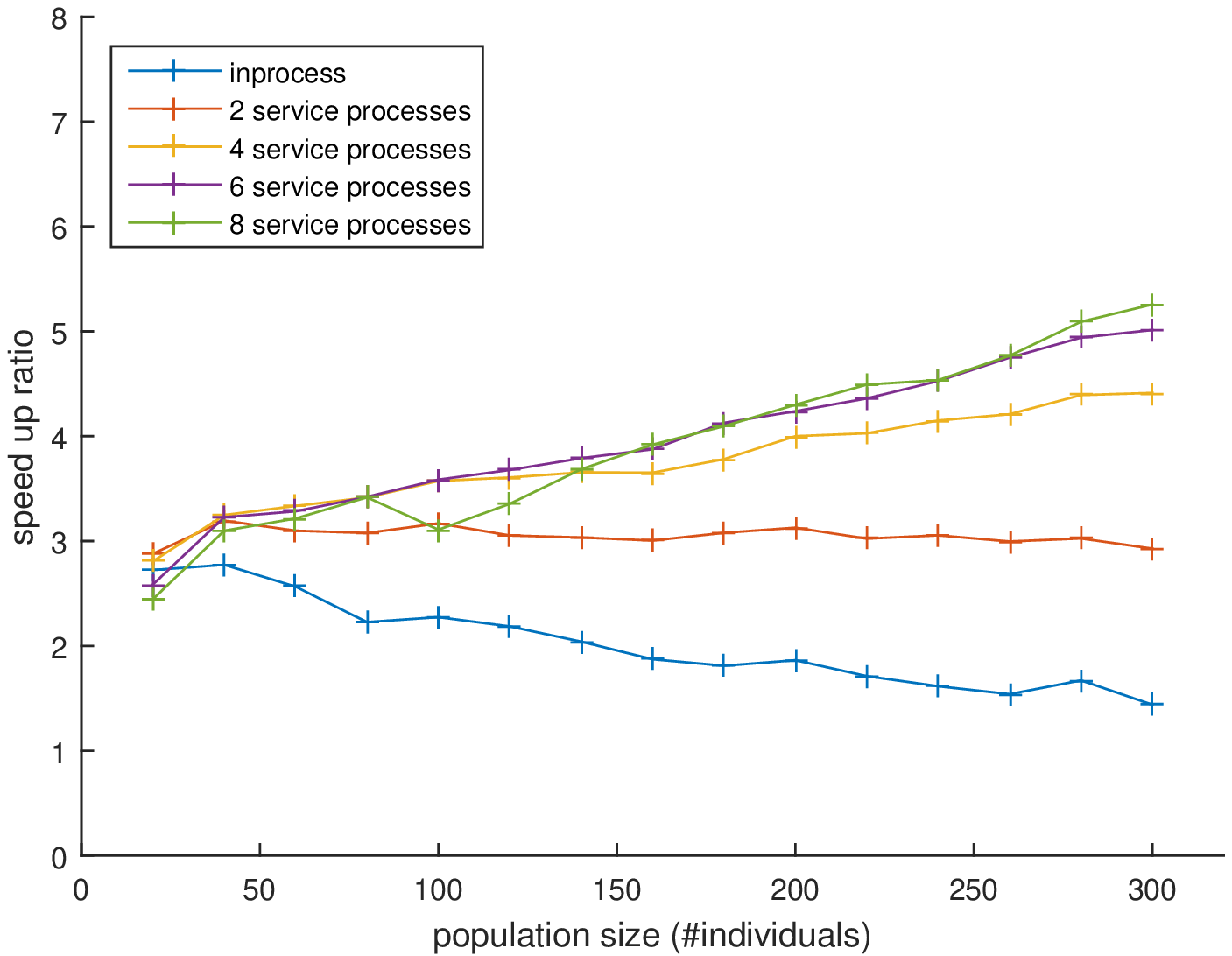}
    \caption{Speedup against conventional compilation }
    \end{subfigure}
\begin{subfigure}[b]{0.5\textwidth}
            \centering
            \includegraphics[width=\textwidth]{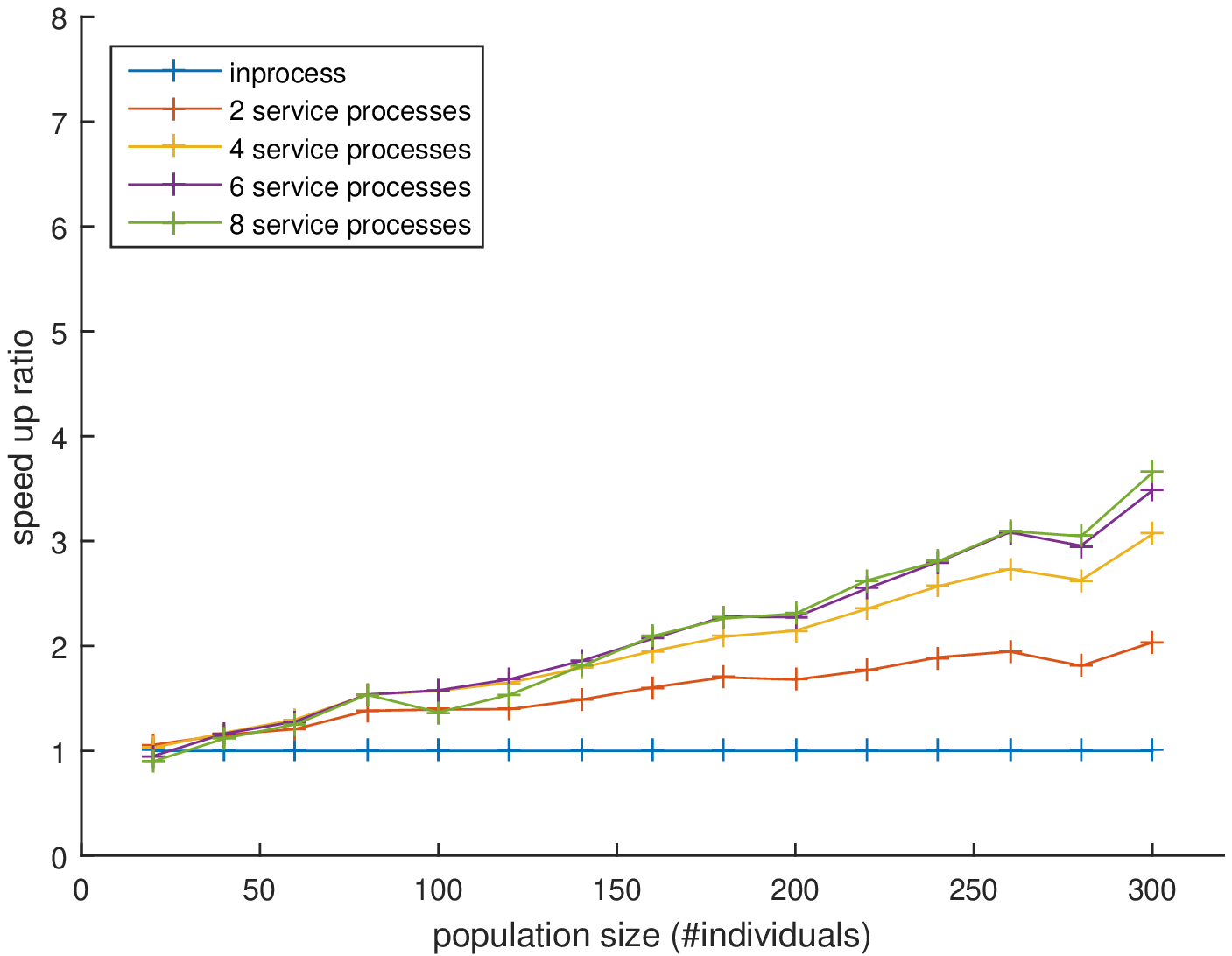}
    \caption{Speedup against in-process compilation}
    \end{subfigure}

\caption{Parallelization speedup on Search problem}
\label{fig:search-speedup-vs-nvrtc}
\end{figure}

\begin{figure}[!ht]  %  ----------------

    \begin{subfigure}[b]{0.5\textwidth}
            \centering
            \includegraphics[width=\textwidth]{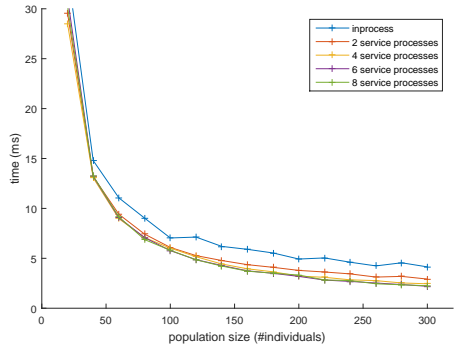}
    \caption{Per individual compile time}
    \end{subfigure}
\begin{subfigure}[b]{0.5\textwidth}
            \centering
            \includegraphics[width=\textwidth]{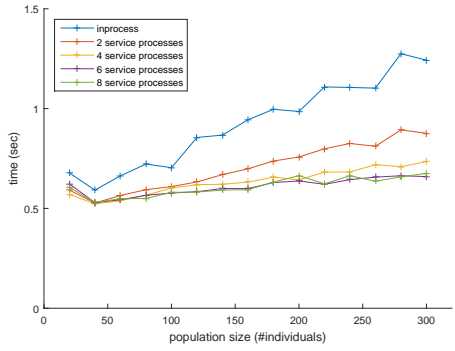}
    \caption{Total compile time}
    \end{subfigure}

\caption{Nvcc compilation times for Keijzer-6 regression by number of servicing resident processes}
\label{fig:K6-parallel}
\end{figure}

\begin{figure}[!ht]  %  ----------------

    \begin{subfigure}[b]{0.5\textwidth}
            \centering
            \includegraphics[width=\textwidth]{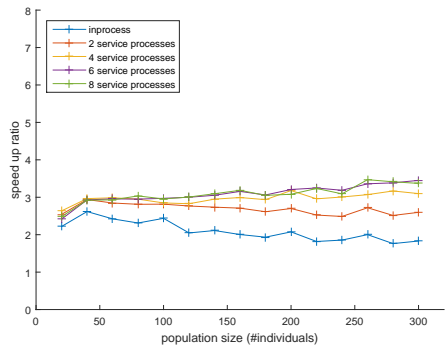}
    \caption{Speedup against conventional compilation }
    \end{subfigure}
\begin{subfigure}[b]{0.5\textwidth}
            \centering
            \includegraphics[width=\textwidth]{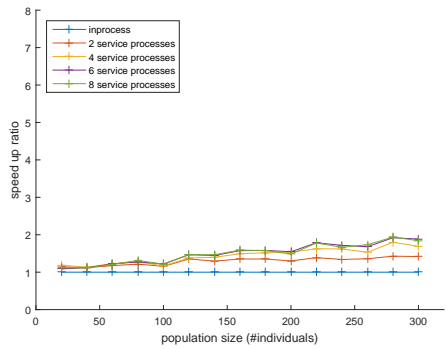}
    \caption{Speedup against in-process compilation}
    \end{subfigure}

\caption{Parallelization speedup on Keijzer-6 regression}
\label{fig:K6-speedup-vs-nvrtc}
\end{figure}

\begin{figure}[!ht]  %  ----------------

    \begin{subfigure}[b]{0.5\textwidth}
            \centering
            \includegraphics[width=\textwidth]{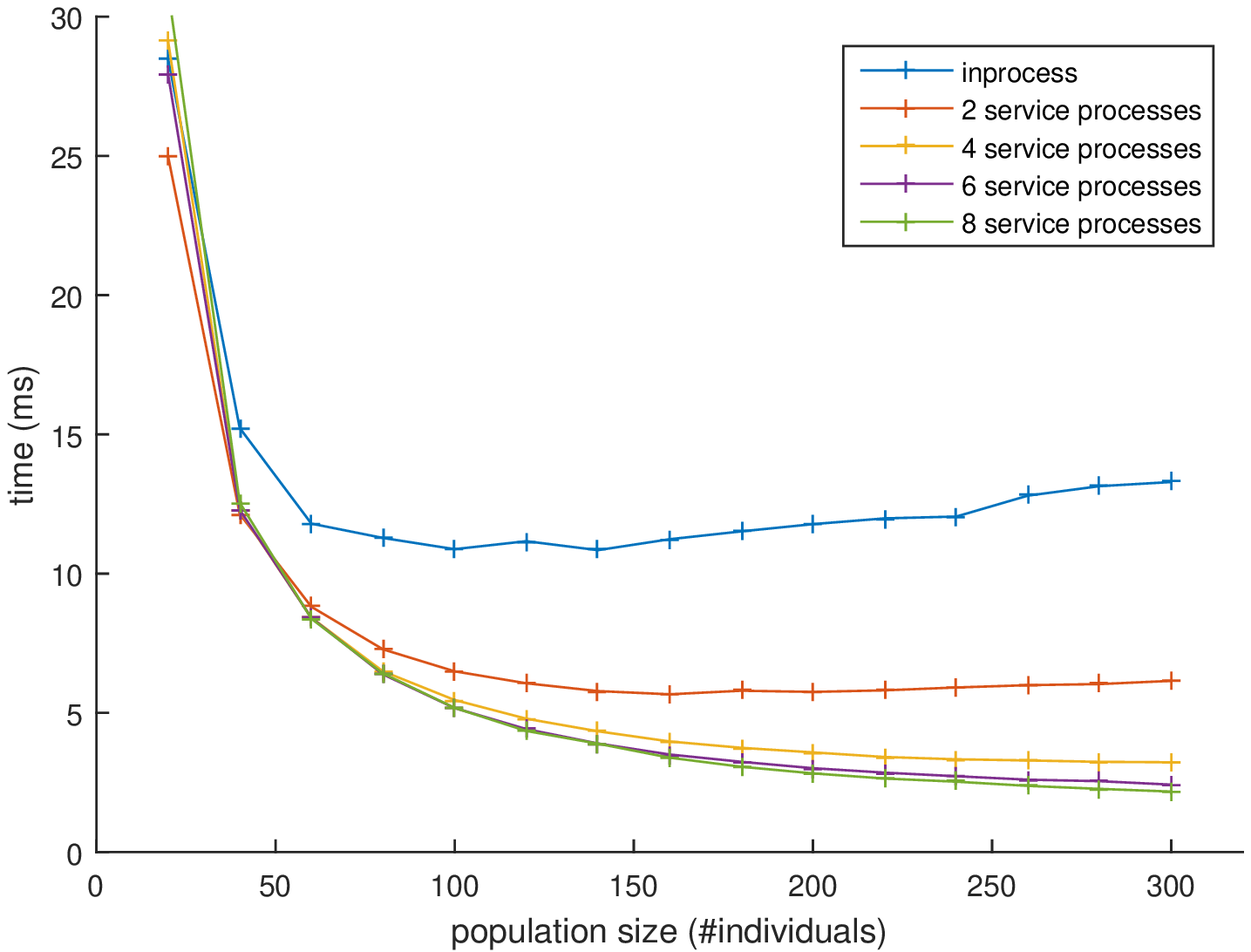}
    \caption{Per individual compile time}
    \end{subfigure}
\begin{subfigure}[b]{0.5\textwidth}
            \centering
            \includegraphics[width=\textwidth]{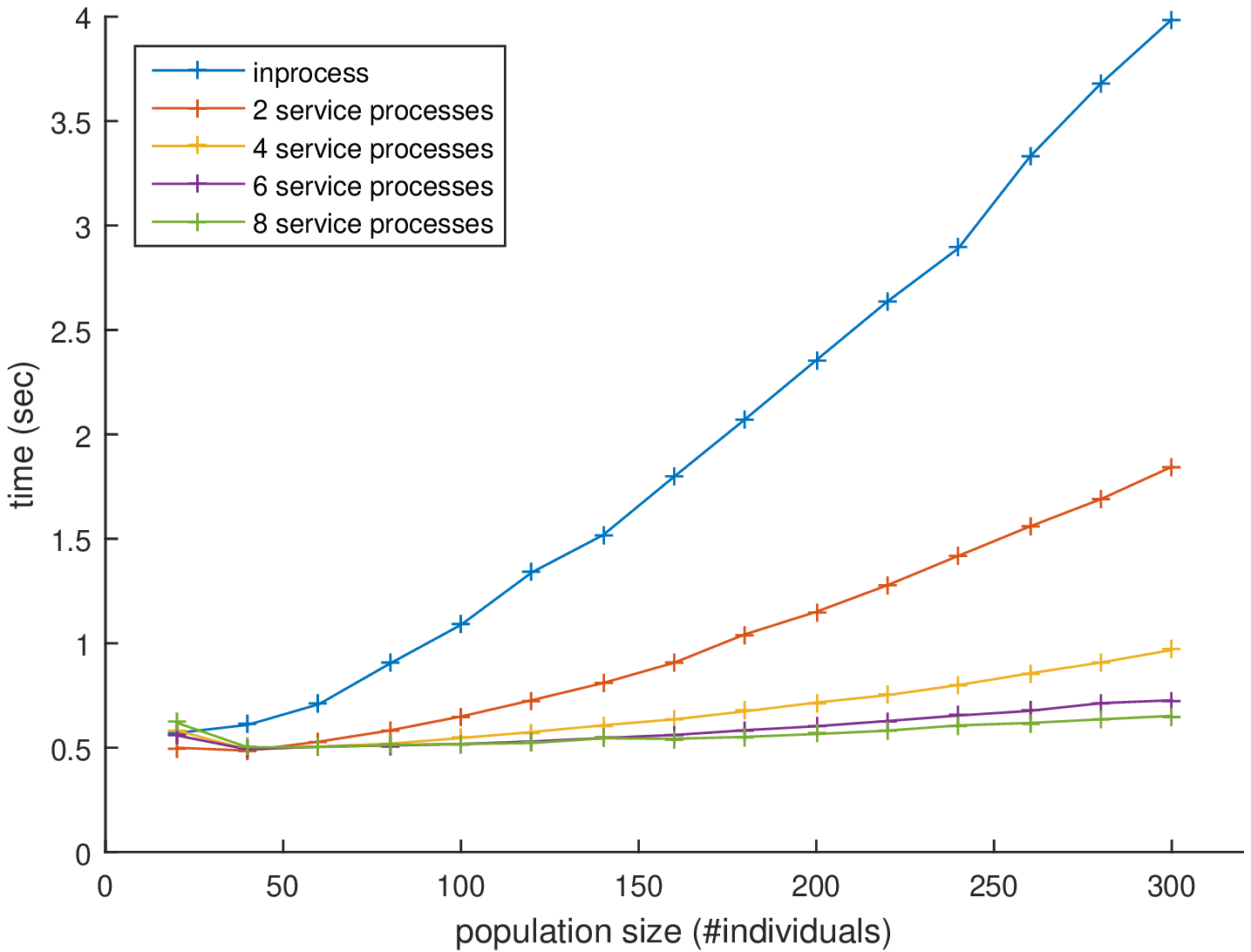}
    \caption{Total compile time}
    \end{subfigure}

\caption{Nvcc compilation times for 5-bit Multiplier by number of servicing resident processes}
\label{fig:MUL-parallel}
\end{figure}

\begin{figure}[!ht]  %  ----------------

    \begin{subfigure}[b]{0.5\textwidth}
            \centering
            \includegraphics[width=\textwidth]{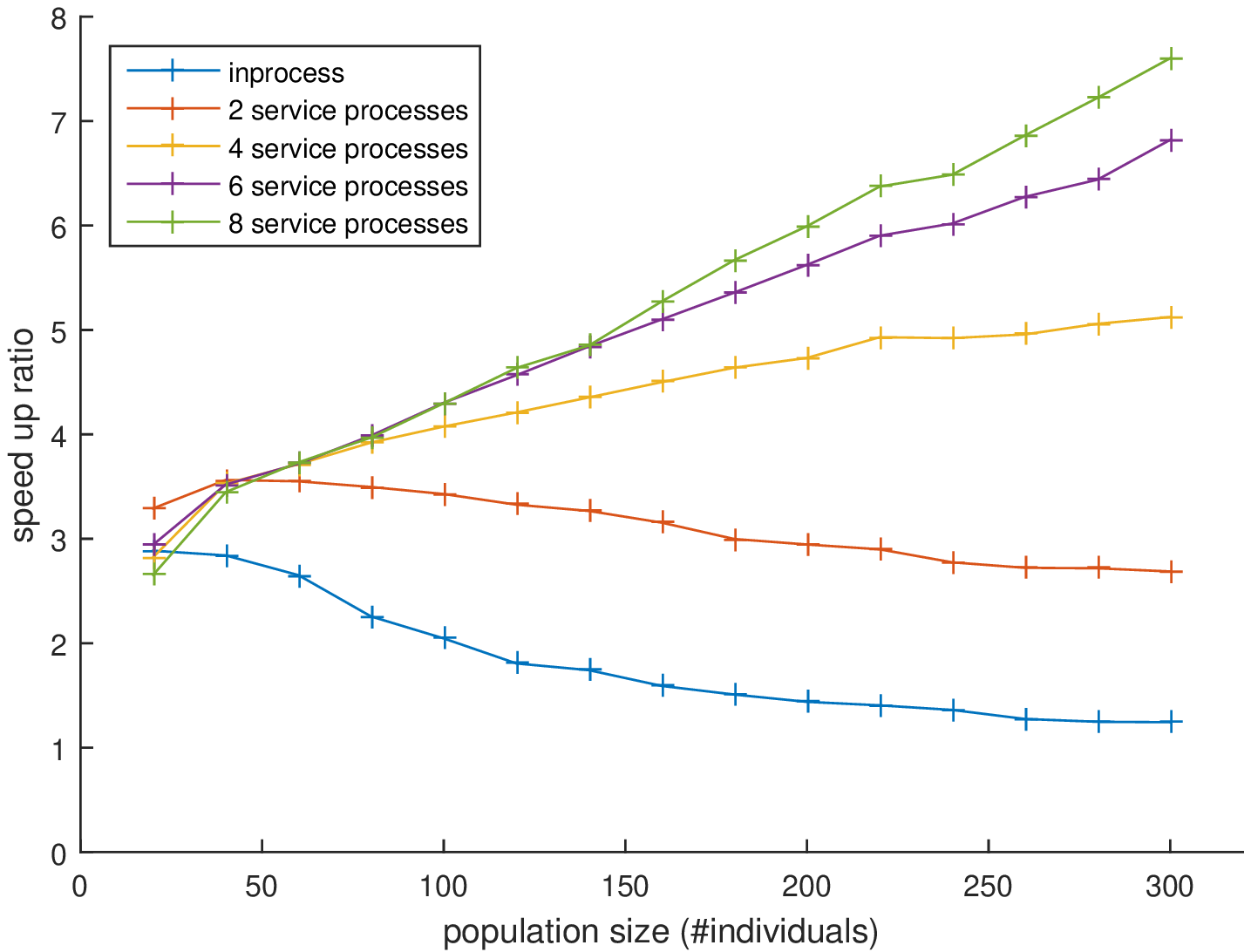}
    \caption{Speedup against conventional compilation }
    \end{subfigure}
\begin{subfigure}[b]{0.5\textwidth}
            \centering
            \includegraphics[width=\textwidth]{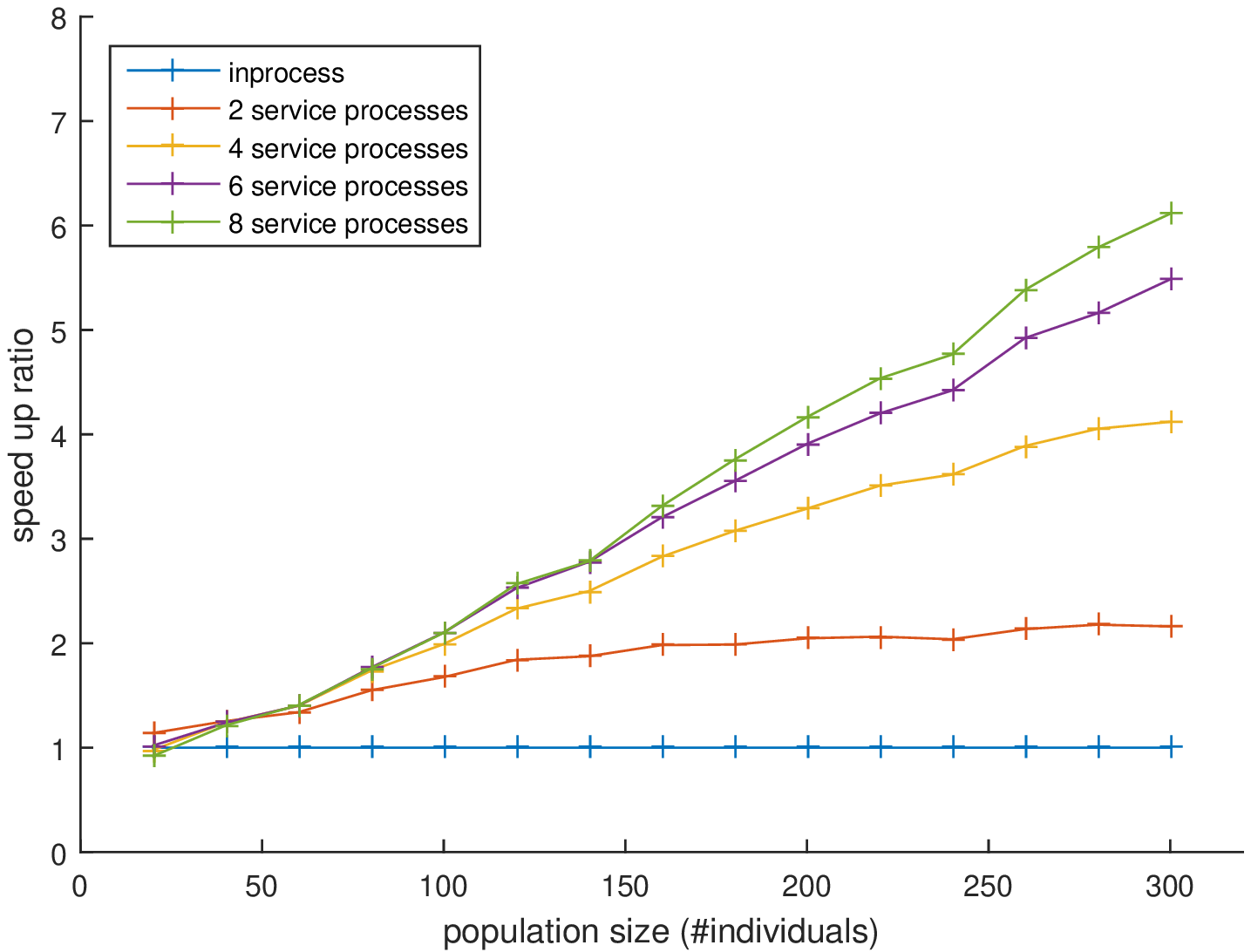}
    \caption{Speedup against in-process compilation}
    \end{subfigure}

\caption{Parallelization speedup on 5-Bit multiplier}
\label{fig:MUL-speedup-vs-nvrtc}
\end{figure}

At the end of each batch of experiments main application dumps the collected raw measurements to a file. We imported this data to Matlab filtered by experiment and measurement types, and aggregated the experiment values for each population size to produce the Tables \ref{search-table},\ref{K6-table},\ref{MUL-table}, and to create the  Figures \ref{fig:search-parallel},\ref{fig:search-speedup-vs-nvrtc},\ref{fig:K6-parallel},\ref{fig:K6-speedup-vs-nvrtc},\ref{fig:MUL-parallel},\ref{fig:MUL-speedup-vs-nvrtc}.

It can be seen that parallelized in-process compilation of genetic programming individuals is faster for all problems and population sizes when compared to in-process compilation without parallelization; furthermore in-process compilation without parallelization itself was shown to be faster than regular command line nvcc compilation on previous section.

Parallel compilation brought the per individual compilation time to 2.17 ms/individual for 5-bit Multiplier, to 2.20 ms/individual for Keijzer-6 regression and to 2.13 milliseconds for the Search Problem; these are almost an order of magnitude faster than previous published results. Also we measured a compilation speedup of $\times 3.45$ for regression problem, $\times 5.26$ for search problem, and $\times 7.60$ for multiplication problem, when compared to the latest Nvcc V8 compiler, without requiring any code modification, and without any runtime performance penalty.

Notice that our experiment platform consisted of dual Xeon E5-2670 processors running at 2.6Ghz; for compute bound tasks increase on processor frequency almost directly translates to performance improvement at an equal rate\footnote{assuming all other things being equal}. Therefore we can conjecture that to be able to compile a population of 300 individuals at sub-millisecond durations, the required processor frequency is around $ 2.6 \times 2.13 = 5.54Ghz$\footnote{once again, under assumption of all other things being equal. 2.13 is the compilation time of Search Problem with 8 daemons} which is currently available.

\section*{Conclusion}

In this paper we present a new method to accelerate the compilation of genetic programming individuals, in order to  keep the compiled approach as a viable option for genetic programming on gpu.

By using an in-process GPU compiler, we replaced disk file based data transfer to/from the compiler with memory accesses, also we mitigated the overhead of repeated launches and tear downs of the command line compiler. Also we investigated ways to parallelize this method of compilation, and identified that in-process compilation function automatically serializes concurrent calls from different threads. We implemented a daemon process that can have multiple running instances and service another application requesting CUDA code compilation. Daemon processes use the same in-line compilation method and communicate through operating system's Inter Process Communication primitives.

We measured compilation times just above 2.1 ms/individual for all three benchmark problems; and observed compilation speedups ranging from $\times 3.45$ to $\times 7.60$ based on problem, when compared to repeated command line compilation with latest Nvcc v8 compiler.

All data and source code of software presented in this paper is available at https://github.com/hayral/Parallel-and-in-process-compilation-of-individuals-for-genetic-programming-on-GPU

\section*{Acknowledgments}
Dedicated to the memory of Professor Ahmet Coşkun Sönmez.
\\First author was partially supported by Turkcell Academy.

\section*{Appendix}

\subsection*{Search Problem}  %  -------------------------------------------------------------
% \subsubsection*{Compilation Time and Speedup Ratio Plots}
% \begin{figure}[!htb]  %  ----------------

%     \begin{subfigure}[b]{0.5\textwidth}
%             \centering
%             \includegraphics[width=\textwidth]{search-parallel.eps}
%     \caption{Per individual compile time}
%     \end{subfigure}
% \begin{subfigure}[b]{0.5\textwidth}
%             \centering
%             \includegraphics[width=\textwidth]{search-parallel-total.eps}
%     \caption{Total compile time}
%     \end{subfigure}

% \caption{Nvcc compilation times for Search Problem by number of servicing resident processes}
% \label{fig:search-parallel}
% \end{figure}

% \begin{figure}[!htb]  %  ----------------

%     \begin{subfigure}[b]{0.5\textwidth}
%             \centering
%             \includegraphics[width=\textwidth]{search-speedup-vs-nvcc.eps}
%     \caption{Speedup ratio against conventional compilation }
%     \end{subfigure}
% \begin{subfigure}[b]{0.5\textwidth}
%             \centering
%             \includegraphics[width=\textwidth]{search-speedup-vs-nvrtc.eps}
%     \caption{Speedup ratio against in-process compilation}
%     \end{subfigure}

% \caption{Parallelization speedup on Search problem}
% \label{fig:search-speedup-vs-nvrtc}
% \end{figure}

\subsubsection*{ Grammar Listing}
\begin{lstlisting}[captionpos=b,caption=Grammar for Search Problem, basicstyle=\footnotesize, belowcaptionskip=4pt ]
<expr>        ::= <expr2> <bi-op> <expr2> | <expr2>
<expr2>       ::= <int> | <var-read> | <var-indexed>

<var-read>    ::= tmp | i | OUTPUT | SEARCH
<var-indexed> ::= INPUT[<var-read> % LENINPUT]
<var-write>   ::= tmp | OUTPUT

<bi-op>       ::= + | -

<int>         ::= 1 | 2 | (-1)

<statement>   ::= <assignment> | <if> | <loop>
<statement2>  ::= <assignment> | <if2>
<statement3>  ::= <assignment>

<loop>        ::= for(i=0;i #lesser# LENINPUT;i++){<c-block2>}

<if>          ::= if (<cond-expr>) {<c-block2>}
<if2>         ::= if (<cond-expr>) {<c-block3>}

<cond-expr>   ::= <expr> <comp-op> <expr>
<comp-op>     ::= #lesser# | #greater# | == | !=
<assignment>  ::= <var-write> = <expr>;

<c-block>     ::= <statements>
<c-block2>    ::= <statements2>
<c-block3>    ::= <statements3>

<statements>  ::= <statement> 
                  | <statement><statement> 
                  | <statement><statement><statement>
                  
<statements2> ::= <statement2> 
                  | <statement2><statement2> 
                  | <statement2><statement2><statement2>
                  
<statements3> ::= <statement3> 
                  | <statement3><statement3> 
                  | <statement3><statement3><statement3>
\end{lstlisting}

\subsubsection*{Code Preamble for Whole Population}
\begin{lstlisting}[captionpos=b,caption=Code preamble for whole population on Search Problem, basicstyle=\footnotesize, belowcaptionskip=4pt ]
__constant__ int _INPUT[NUMTESTCASE][MAX_TESTCASE_LEN];
__constant__ int _LENINPUT[NUMTESTCASE];
__constant__ int _SEARCH[NUMTESTCASE];
__constant__ int CORRECTANSWER[NUMTESTCASE];


__global__ void createdFunc(int *_OUTPUT)
{
    int *INPUT = _INPUT[threadIdx.x];
    int LENINPUT = _LENINPUT[threadIdx.x];
    int SEARCH = _SEARCH[threadIdx.x];
    
    int i;
    int tmp;
    int OUTPUT;
\end{lstlisting}

\subsection*{Keijzer-6 Regression Problem}  %  -------------------------------------------------------------
% \subsubsection*{Compilation Time and Speedup Ratio Plots}

% \begin{figure}[!htb]  %  ----------------

%     \begin{subfigure}[b]{0.5\textwidth}
%             \centering
%             \includegraphics[width=\textwidth]{K6-parallel.eps}
%     \caption{Per individual compile time}
%     \end{subfigure}
% \begin{subfigure}[b]{0.5\textwidth}
%             \centering
%             \includegraphics[width=\textwidth]{K6-parallel-total.eps}
%     \caption{Total compile time}
%     \end{subfigure}

% \caption{Nvcc compilation times for Keijzer-6 regression by number of servicing resident processes}
% \label{fig:K6-parallel}
% \end{figure}

% \begin{figure}[!htb]  %  ----------------

%     \begin{subfigure}[b]{0.5\textwidth}
%             \centering
%             \includegraphics[width=\textwidth]{K6-speedup-vs-nvcc.eps}
%     \caption{Speedup ratio against conventional compilation }
%     \end{subfigure}
% \begin{subfigure}[b]{0.5\textwidth}
%             \centering
%             \includegraphics[width=\textwidth]{K6-speedup-vs-nvrtc.eps}
%     \caption{Speedup ratio against in-process compilation}
%     \end{subfigure}

% \caption{Parallelization speedup on Keijzer-6 regression}
% \label{fig:K6-speedup-vs-nvrtc}
% \end{figure}

\subsubsection*{ Grammar Listing}
\begin{lstlisting}[captionpos=b,caption=Grammar for Keijzer-6 Regression, basicstyle=\footnotesize, belowcaptionskip=4pt ]
<e>  ::= <e2> + <e2> | <e2> - <e2> | <e2> * <e2> | <e2> / <e2>
         | sqrtf(fabsf(<e2>)) | sinf(<e2>) | tanhf(<e2>)
         | expf(<e2>) | logf(fabsf(<e2>)+1)
         | x | x | x | x
         | <c><c>.<c><c> | <c><c>.<c><c> | <c><c>.<c><c> | <c><c>.<c><c>
         
<e2> ::= <e3> + <e3> | <e3> - <e3> | <e3> * <e3> | <e3> / <e3>
         | sqrtf(fabsf(<e3>)) | sinf(<e3>) | tanhf(<e3>)
         | expf(<e3>) | logf(fabsf(<e3>)+1)
         | x | x | x | x | x | x
         | <c><c>.<c><c> | <c><c>.<c><c> | <c><c>.<c><c>
         | <c><c>.<c><c> | <c><c>.<c><c> | <c><c>.<c><c>
         
<e3> ::= <e3> + <e3> | <e3> - <e3> | <e3> * <e3> | <e3> / <e3>
         | sqrtf(fabsf(<e3>)) | sinf(<e3>) | tanhf(<e3>)
         | expf(<e3>) | logf(fabsf(<e3>)+1)
         | x | x | x | x | x | x | x | x
         | <c><c>.<c><c> | <c><c>.<c><c> | <c><c>.<c><c>
         | <c><c>.<c><c> | <c><c>.<c><c> | <c><c>.<c><c>
         | <c><c>.<c><c> | <c><c>.<c><c>
         
<c>  ::= 0|1|2|3|4|5|6|7|8|9
\end{lstlisting}

\subsection*{5-bit Multiplier Problem}  %  -------------------------------------------------------------
% \subsubsection*{Compilation Time and Speedup Ratio Plots}

% \begin{figure}[!htb]  %  ----------------

%     \begin{subfigure}[b]{0.5\textwidth}
%             \centering
%             \includegraphics[width=\textwidth]{MUL-parallel.eps}
%     \caption{Per individual compile time}
%     \end{subfigure}
% \begin{subfigure}[b]{0.5\textwidth}
%             \centering
%             \includegraphics[width=\textwidth]{MUL-parallel-total.eps}
%     \caption{Total compile time}
%     \end{subfigure}

% \caption{Nvcc compilation times for 5-bit Multiplier by number of servicing resident processes}
% \label{fig:MUL-parallel}
% \end{figure}

% \begin{figure}[!htb]  %  ----------------

%     \begin{subfigure}[b]{0.5\textwidth}
%             \centering
%             \includegraphics[width=\textwidth]{mul-speedup-vs-nvcc.eps}
%     \caption{Speedup ratio against conventional compilation }
%     \end{subfigure}
% \begin{subfigure}[b]{0.5\textwidth}
%             \centering
%             \includegraphics[width=\textwidth]{mul-speedup-vs-nvrtc.eps}
%     \caption{Speedup ratio against in-process compilation}
%     \end{subfigure}

% \caption{Parallelization speedup on 5-Bit multiplier}
% \label{fig:MUL-speedup-vs-nvrtc}
% \end{figure}

\subsubsection*{ Grammar Listing}
\begin{lstlisting}[captionpos=b,caption=Grammar for 5-bit Multiplier Problem, basicstyle=\footnotesize, belowcaptionskip=4pt ]
<start> ::= o0=<expr>;o1=<expr>;o2=<expr>;o3=<expr>;o4=<expr>;
            o5=<expr>;o6=<expr>;o7=<expr>;o8=<expr>;o9=<expr>;
<expr>  ::= (<expr2> <bi-op> <expr2>) | <var> | (~ <var>)
<expr2> ::= (<expr2> <bi-op> <expr2>) | <var> | (~ <var>)
            | <var> | (~ <var>)
<var>   ::= a0 | a1 | a2 | a3 | a4 | b0 | b1 | b2 | b3 | b4
<bi-op> ::= & | #or#
\end{lstlisting}
\subsubsection*{Code Preamble for each Individual}
\begin{lstlisting}[captionpos=b,caption=Code preamble for 5-bit Multiplier Problem, basicstyle=\footnotesize, belowcaptionskip=4pt ]
__global__ void createdFunc0(int *_OUTPUT)
{
int tid = blockIdx.x *blockDim.x + threadIdx.x;;

int a0 = tid & 0x1;
int a1 = (tid & 0x2) >> 1;
int a2 = (tid & 0x4) >> 2;
int a3 = (tid & 0x8) >> 3;
int a4 = (tid & 0x10) >> 4;
int b0 = (tid & 0x20) >> 5;
int b1 = (tid & 0x40) >> 6;
int b2 = (tid & 0x80) >> 7;
int b3 = (tid & 0x100) >> 8;
int b4 = (tid & 0x200) >> 9;

int o0,o1,o2,o3,o4,o5,o6,o7,o8,o9;
\end{lstlisting}

\subsection*{Compilation Time and Speedup Ratio Plots}

\clearpage

\bibliographystyle{plain} 
\bibliography{sample}

\end{document}